\documentclass[11pt]{article}

\usepackage[a4paper, margin=1in]{geometry} 
\usepackage[utf8]{inputenc}
\usepackage[T1]{fontenc}
\usepackage{lmodern}
\usepackage{graphicx}
\usepackage{hyperref}
\usepackage{amsmath,amssymb}
\usepackage{bm}
\usepackage{verbatim}
\usepackage{algorithm} 
\usepackage{algpseudocode} 
\usepackage{xcolor}

\newcommand{\beq}{\begin{equation}}
\newcommand{\eeq}{\end{equation}}

\title{Generalized Kalman Filter based Temporal Difference Reinforcement Learning }

\author{
\begin{tabular}{p{0.9\textwidth}}
    \centering
    Vasos Arnaoutis $^{1,*}$ \quad
     Eric Lutters $^{2}$ \quad
     Bojana Rosi\'c $^{1,3}$ \\
     {\small v.arnaoutis@utwente.nl \quad d.lutters@utwente.nl \quad bojana.rosic@tuwien.ac.at}\\
    \flushleft
    {\small
    $^{1}$ Applied Mechanics and Data Analysis, Faculty of Engineering Technology, University of Twente, Enschede, the Netherlands \\
    $^{2}$ Department of Design, Production and Management, Faculty of Engineering Technology, University of Twente, Enschede, the Netherlands \\
    $^{3}$ Digital Engineering research group, Faculty of Mechanical and Industrial Engineering, Vienna University of Technology, Vienna, Austria \\
    * Correspondence: v.arnaoutis@utwente.nl
    }
\end{tabular}
}

\date{\today}

\begin{document}

\maketitle

\begin{abstract}
\noindent In this paper, we present a generalized temporal-difference (TD) reinforcement learning framework based on the theory of conditional expectations. The value and action-value (Q-value) functions are treated as uncertain quantities, and their estimation is formulated as a stochastic inference problem. Unlike classical Kalman-based temporal-difference learning, which relies on linear-Gaussian assumptions, the proposed formulation is derived directly from the conditional expectation framework and naturally extends to nonlinear models and non-Gaussian probability distributions.
The proposed method recursively estimates not only the conditional expectation of the value function but also its second probabilistic moment, thereby quantifying the uncertainty associated with the learned value function throughout the learning process. To obtain a computationally tractable algorithm, the stochastic problem is discretized using either polynomial chaos expansions or ensemble-based approximations, providing efficient representations of the underlying random variables.
The proposed framework is demonstrated on two optimal control problems: a linear mass--spring--damper system and a nonlinear heat conduction problem in a closed cavity. The numerical examples illustrate the capability of the proposed method to accurately estimate both the value function and its associated uncertainty, while extending classical Kalman-based temporal-difference learning to a broader class of stochastic systems.
\end{abstract}


\newcommand{\E}{\mathbb{E}} 
\newcommand{\N}{\mathcal{N}} 
\newcommand{\U}{\mathcal{U}} 
\newcommand{\R}{\mathbb{R}} 
\newcommand{\pdf}{p} 
\newcommand{\Salg}{\mathfrak{F}} 
\newcommand{\Ssalg}{\mathfrak{B}} 
\newcommand{\Prob}{\mathbb{P}} 
\newcommand{\V}{V} 
\newcommand{\Proj}{P}
\newcommand{\Space}{\mathcal{S}}
\section{Introduction}

Modern computational tools leverage both model-based and data-driven strategies for the control and design optimization of mechanical systems, which are typically formulated as optimization problems \cite{Kim2025AutomaticLearning, Berry2016SemiparametricModels}. As these systems are often nonlinear and complex, data-driven methods have emerged as promising approaches for tackling the associated challenges. Among them, model-free reinforcement learning has become one of the leading techniques \cite{Wiering2012}. The core idea is to represent the system as a Markov decision process (MDP) and to design an agent that learns an actuation policy aimed at maximizing the cumulative return over time \cite{Sutton1998, Hayashi2020, Jang2022}, thus leading to optimal control or design. One of the earliest model-free reinforcement learning algorithms is the temporal difference (TD) learning \cite{sutton1988td} that has been particularly effective in solving problems that are of discrete and low-dimensional nature. When combined with function approximation, TD learning can also be extended to continuous and high-dimensional problems \cite{Sutton1998}. Prominent examples of such extensions include Deep Q-Networks (DQN), Deep Deterministic Policy Gradient (DDPG), and Soft Actor–Critic (SAC) \cite{Mnih2013PlayingLearning,Lillicrap2015, Haarnoja2018, Nguyen2021}. When viewed from a Bayesian perspective, the TD algorithm can be interpreted as a form of Kalman filtering, as demonstrated in \cite{Geist2010,Geist2013, Baird1995}. This formulation enables faster convergence through regularization, where prior knowledge is incorporated as a probabilistic constraint.

Building on this probabilistic formulation, several multi-modal extensions of KTD have been proposed. These methods are particularly studied in the context of successor representation temporal difference learning \cite{Geerts2019, Malekzadeh2020, Salimibeni2020}.
A closely related line of work adopts a stochastic process perspective on temporal difference learning, most notably Gaussian Process Temporal Difference (GPTD) learning \cite{Engel2003, Engel2005}, which models the value function as a Gaussian process. To increase representational capacity beyond a single kernel, ensemble-based variants of GPTD introduce weighted mixtures of Gaussian priors \cite{Lu2021}.

From a different perspective, value function learning can be interpreted as a projection problem rather than an incremental update scheme. In this formulation, complementary to temporal-difference-based methods, a class of model-free approaches relies on projection operators that enforce Bellman consistency by projecting the Bellman update onto a chosen function space \cite{Sutton1998}. The learning problem is thereby cast as finding the closest approximation of the Bellman target within a restricted hypothesis class, effectively reformulating value function learning as a sequence of constrained approximation steps.
A canonical example is Least Squares Temporal Difference (LSTD) learning \cite{Bradtke1996}, which computes the value function directly by minimizing the squared error within a restricted function space. This idea was later extended to control settings through Least Squares Policy Iteration (LSPI), which embeds the same projection principle within a policy iteration framework \cite{Lagoudakis2003}.

From a Bayesian reinforcement learning perspective, these least-squares approaches can also be interpreted as performing approximate posterior inference over the value function. In this interpretation, the projection step corresponds to incorporating observations into a regularized inference update, linking least-squares solutions to Bayesian updating under Gaussian assumptions \cite{Tziortziotis2017BayesianRegularization}.

From the projection viewpoint, the previously discussed approaches can be further generalized within a broader reinforcement learning framework by interpreting conditional expectation as an orthogonal projection \cite{Luenberger1968}. In probability theory, this means that the best estimate of a random variable, given the available information, is obtained by projecting it onto the space of all functions consistent with that information \cite{Matthies2016}. The resulting projection is characterized by minimizing the mean-squared distance between the true quantity and its approximation, making it the optimal estimate in the $L^2$ sense.
When the set of admissible functions is restricted to a parametric class, this projection becomes equivalent to solving a least-squares estimation problem. This observation provides a unifying perspective on temporal difference learning and least-squares methods, as both can be interpreted as different computational realizations of the same underlying projection principle.

According to the Doob–Dynkin lemma \cite{Bobrowski2005}, the conditional expectation of the return given the current state can be represented as a measurable function of the available information or observations. When this function is restricted to a linear parameterization, the resulting formulation recovers classical least-squares estimation and Kalman filter-based reinforcement learning methods. In this work, we extend this viewpoint beyond linear settings by considering nonlinear function classes and more general discrepancy measures, leading to a unified Gauss–Markov–Kalman filter temporal difference (GMKF-TD) learning framework. In contrast to existing approaches that primarily estimate only the conditional mean, our formulation explicitly estimates the conditional covariance, thereby enabling richer uncertainty quantification in nonlinear and non-Gaussian regimes. Within this framework, the value function is interpreted as an uncertain quantity, reflecting epistemic uncertainty about its true form. This uncertainty is consistently transferred to a parameterized representation, where the parameters are treated as random variables. For practical implementation, these random variables are represented either through Monte Carlo sampling or functional approximation techniques such as polynomial chaos expansion (PCE) \cite{Bojana2012}, depending on the desired trade-off between expressiveness and computational efficiency.

This viewpoint naturally connects to a range of existing approaches that study how such uncertainty can be represented and propagated in reinforcement learning. In particular, the PCE approximation has been used to construct functional representations of the value function over state spaces in model-based reinforcement learning approaches, enabling near-optimal policy computation in structured settings \cite{Schweitzer1985GeneralizedProcesses}. In \cite{Liu2025Model-freeChaos} a model-free robust reinforcement learning method for continuous control based on robust Markov decision processes is proposed. Here, uncertainty is modeled via stochastic parameterized uncertainty sets and the robust value function is estimated using generalized PCE within an off-policy actor–critic framework, with convergence guarantees and improved robustness over standard methods. Furthermore, in \cite{Zarrouki2024AdaptiveControl} PCE is used to represent and propagate uncertainty in the system dynamics by expressing them in a polynomial basis, allowing the learning algorithm to efficiently account for stochastic effects during policy optimization in continuous control setting.

Building on these ideas, we adopt PCE representations within our framework as an efficient and flexible strategy for modeling the random parameters of the value function. Furthermore, we reformulate temporal difference learning as a generalized Kalman filtering algorithm based on this PCE approximation.

The remainder of the paper is organized as follows. We first provide a brief introduction to reinforcement learning and Bayesian inference, which serves as the basis for deriving the proposed GMKF-TD framework. We then analyze the linear case of GMKF-TD in its probabilistic form. For stochastic representations, we consider two discretization strategies: PCE and ensemble-based approximations. The method is further extended from state-value functions to action-value functions, followed by the introduction of a variance-based exploration strategy compatible with the proposed framework. Finally, numerical results are presented on two physics-based benchmarks, namely a controllable mass–spring–damper system and a closed cavity heat transfer problem.

\section{Problem definition}

Let us consider a physical system whose dynamics are described by the 
parameter-dependent evolution equation
\begin{equation}
\label{evequation}
\frac{d}{dt} y(t;p)
+ A\big(y(t;p);p\big)
+ W\big(y(t;p);p\big)
= F\big(y(t;p);p\big),
\end{equation}
for $t \in (0,T]$, subject to the initial condition
\begin{equation}
y(0) = y_0 \in \mathcal{H}.
\end{equation}
Here, $y(t;p) \in \mathcal{Y}$ denotes the system state, with 
$\mathcal{Y}$ being a separable Hilbert space, and 
$p \in \mathcal{P}$ represents parameters that may characterize 
material properties or geometric features of the system, or, more generally, 
design variables of the system. The operator $A : \mathcal{Y} \to \mathcal{Y}^{*}$ ($\mathcal{Y}^{*}$ being the dual of $\mathcal{Y}$) represents the linear
(dissipative) part of the dynamics and is assumed to be bounded and
coercive in the sense that there exists $\alpha > 0$ such that
\[
\langle A(y), y \rangle_{\mathcal{Y}^{*},\mathcal{Y}}
\geq \alpha \|y\|_{\mathcal{Y}}^{2}
\quad \text{for all } y \in \mathcal{Y}.
\]
This operator typically models diffusive or elliptic effects. On the other hand, the mapping 
$W : \mathcal{Y} \to \mathcal{Y}^{*}$ denotes the nonlinear
contribution. We assume that $W$ is locally Lipschitz
continuous from $\mathcal{Y}$ into $\mathcal{Y}^{*}$, and satisfies a suitable
skew-symmetry or energy-balance condition, for instance
\[
\langle W(y), y \rangle_{\mathcal{Y}^{*},\mathcal{Y}} = 0
\quad \text{for all } y \in \mathcal{Y},
\]
which guarantees that it does not contribute directly to the
dissipation of energy.
The operator 
$F : \mathcal{Y} \to \mathcal{Y}^{*}$ collects coupling
effects and external forcing terms. In general, $F$
may depend nonlinearly on $y$ and $p$.

This abstract setting provides a unified framework for a
broad class of coupled nonlinear evolution systems.
In subsequent sections, the concrete physical model
under consideration will be identified as a particular
realization of the operators $A$, $W$, and $F$
within this general structure.

The goal is to design or control the system through a set of actions $p,u$ assuming that the dynamics of the system are unknown, and have to be identified. In other words, we define a performance functional
\begin{equation}
J(p,u) =
\int_0^T \ell(y(t;p,u),u(t),p)\, dt
\, + \ell_T(y(T;p,u),p),
\end{equation}
where
\[
\ell : \mathcal{Y} \times \mathbb{R}^m \times \mathcal{P} \to \mathbb{R}
\]
is a running cost and
\[
\ell_T : \mathcal{Y} \times \mathcal{P} \to \mathbb{R}
\]
is a terminal cost. Hence, the design and/or control problem reads:
\begin{equation}
(p^*,u^*)=\arg \min_{(p,u)} J(p,u)
\end{equation}
which further has to be solved.
This problem defines an optimization problem in which the state $y$ depends implicitly on $(p,u)$ through the unknown evolution equation. Assuming that the problem is discretized, and a state has a finite dimension, we further interpret the discretized space $\mathcal{Y}_n$ as a state space in which states $S$ take value. Furthermore, we define the space $A$ of all actions (control or design parameter changes), and the corresponding environement transition operator
\[
\mathcal{T} : S\times A
\to S,
\]
implicitly determined by the solution operator of the evolution equation in Eq.~(\ref{evequation}). The goal is then to employ the model-free reinforcement learning algorithm to solve the previous problem by considering the instantaneous reward to be defined as the running cost $\ell(\cdot)$ of the original optimization problem, whereas terminating reward to be related to the terminal cost $\ell_T(\cdot)$. 

The system under consideration is governed by nonlinear partial differential equations, leading to a high-dimensional, time-dependent, and stochastic dynamical system after discretization. While physics-based models provide a consistent description of the underlying processes, the associated optimization and control problems are generally intractable using classical model-based approaches, due to strong nonlinearities, high dimensionality, and the absence of a closed-form solution for the value function. In this context, the problem can be naturally reformulated as a sequential decision process over the evolving system state. We therefore adopt a model-free reinforcement learning approach based on temporal-difference learning, which enables the approximation of value functions directly from simulated trajectories without requiring an explicit solution of the underlying optimal control problem. Although this approach introduces its own assumptions, such as the need for sufficient exploration and stable learning dynamics, it avoids the reliance on explicit model reduction or adjoint-based formulations. The governing PDE model is retained to generate physically consistent trajectories, ensuring that the learned policies remain grounded in the underlying physics while enabling efficient optimization in a complex stochastic setting.

\section{Temporal Difference Learning (TDL)}

In model-free TDL, an agent interacts with an unknown environment to find a series of actions, described by a policy, which maximizes the cumulated rewards over a given horizon. The environment is described by a Markov Decision Process (MDP) \cite{Sutton1998,Puterman2008}, denoted by a tuple $\{ S, A, R, \gamma \}$, in which $S$ and $A$ are the state and action sets, respectively, $R$ is the reward function that assigns the reward received after each state transition, typically depending on the current state and action taken, and $\gamma$ is the discount factor used to diminish the impact of rewards collected far in the future. For an MDP, the probability of a state transition depends only on the agent's current state, also referred to as Markov property. The objective of the agent is to find a policy $\pi: S \times A \mapsto [0,1]$, i.e.~\[
\pi(a|s) = \mathbb{P}( a | s)
\]  that maximizes the return
\begin{equation}
    \label{eqn:G_t}
   G_t = \sum_{k=0}^{\infty} \gamma^k R_{t+k} 
\end{equation}
described by the discounted cumulative rewards over the infinite horizon. 
Given a policy \(\pi\), the expected return starting from state \(s\) and following policy $\pi$, i.e.~the state value function, is defined as
\begin{eqnarray}
    \label{eqn:V_s}
    V_\pi(s) = \E_\pi[G_t |S = s], \quad \forall s \in S.
\end{eqnarray}
 Similarly, one can introduce the state-action value function $Q_\pi(s,a)$ as the expected cumulative discounted reward obtained by starting in state $s$, taking action $a$ immediately, and thereafter following policy $\pi$:
\begin{eqnarray}
    Q_\pi(s,a) = \mathbb{E}_\pi [G_t|S=s, A = a] , \quad \forall s,a \in S \times A.
\end{eqnarray}
The last two quantities are related by:
\[
V_\pi(s) = \sum_{a \in \mathcal{A}} \pi(a|s) \, Q_\pi(s, a),
\]
which expresses the state-value function as the expected value of the action-value function weighted by the policy’s probability of selecting each action.

The value function can be estimated using various methods, which broadly fall into two categories: deterministic \cite{Bradtke1996} and probabilistic approaches \cite{Geist2013, Baird1995}. Unlike deterministic methods that provide only point estimates, probabilistic approaches quantify not only the value function itself but also the uncertainty associated with the estimate. This uncertainty is induced by modeling and measurement errors, as well as by limited data availability, thereby limiting the accuracy of the resulting value function. To quantify and propagate this uncertainty, we develop a framework for estimating its first two moments—namely, the mean and covariance, in both linear and nonlinear problems. The proposed approach is based on projection type of algorithm, motivated by the need to reduce the dimensionality of the problem and enable efficient numerical approximation. In this context, conditional expectation is interpreted as an orthogonal projection onto subspaces associated with partial information, yielding a tractable formulation of the corresponding moment equations.

\section{Projection Method via Conditional Expectations}

In Eq.~(\ref{eqn:V_s}) the expected return conditioned on a specific event (i.e., \( S = s \)) is a deterministic quantity. However, its value is unknown and can therefore be modeled as a random variable. Let \( V_\pi(s) \in L_2(\Omega, \mathcal{F}, \mathbb{P}) \) denote a square-integrable (finite variance) random variable in a probability space \( (\Omega, \mathcal{F}, \mathbb{P}) \), representing the value function under policy \( \pi \) in state \( S = s \). To formalize this within the framework of Hilbert spaces, we interpret conditional expectation as an orthogonal projection onto the closed subspace \( L_2(\Omega,\mathcal{B}, \mathbb{P}) \subseteq L_2(\Omega,\mathcal{F}, \mathbb{P}) \), where \( \mathcal{B} \subseteq \mathcal{F}\) is a sub-\( \sigma \)-algebra representing the available information (e.g., generated by \( S \)). By the orthogonal decomposition theorem~\cite{Kolmogorov1956, Bobrowski2005}, we can write:
\begin{equation}
    \label{eq:orth_decomp}
    V_\pi(s) = \Proj_{\mathcal{B}} V_\pi(s) + (I - \Proj_{\mathcal{B}}) V_\pi(s),
\end{equation}
where \( \Proj_{\mathcal{B}}: L_2(\Omega,\mathcal{F}, \mathbb{P}) \to L_2(\Omega,\mathcal{B}, \mathbb{P}) \) denotes the orthogonal projection operator onto \( L_2(\Omega,\mathcal{B}, \mathbb{P})\). This projection corresponds to the conditional expectation:
\begin{equation}
    \label{eq:cond_exp_proj}
    \mathbb{E}[V_\pi(s) \mid \mathcal{B}] = \Proj_{\mathcal{B}} V_\pi(s),
\end{equation}
and the residual component in Eq.~(\ref{eq:orth_decomp}) is then defined as:
\begin{equation}
    \label{eq:residual}
V_\pi^{\perp}(s):= (I - \Proj_{\mathcal{B}}) V_\pi(s).
\end{equation}
Being orthogonal to all \( \mathcal{B} \)-measurable functions, the residual satisfies
\begin{equation}
    \label{eq:residual_expectation}
    \mathbb{E}[V_\pi^{\perp}(s)\mid \mathcal{B}] = 0.
\end{equation}
By the Doob--Dynkin lemma~\cite{Bobrowski2005}, since \( \Proj_{\mathcal{B}} V_\pi(s) \) is \( \mathcal{B} \)-measurable, there exists a measurable function \( \varphi \) such that:
\begin{equation}
    \label{eq:doob_dynkin}
    \mathbb{E}[V_\pi(s, \omega) \mid \mathcal{B}] = \varphi(s)
\end{equation}
holds. 
Substituting the previous equation into Eq.~\eqref{eq:orth_decomp}, we obtain the decomposition:
\begin{equation}
    \label{eq:decomposition_final}
    V_\pi(s) = \varphi(s) + (V_\pi(s) - \varphi(s)),
\end{equation}
where \( \varphi(s) \) represents the best approximation of \( V_\pi \) given knowledge of the state \( s \), and the second term captures the irreducible uncertainty. Being orthogonal projection, the map $\varphi(\cdot)$, i.e.~\(\hat{V}_\pi:=\varphi(s)\), is the unique element in \(L_2(\Omega,\mathcal{B}, \mathbb{P})\) that minimizes the mean squared error:
\begin{equation}
    \label{eq:mse_minimizer}
    \mathbb{E}[V_\pi \mid \mathcal{B}] := \Proj_{\mathcal{B}} V_\pi = \arg\min_{\hat{V}_\pi \in L_2(\mathcal{B})} \left\| V_\pi - \hat{V}_\pi \right\|^2_{L_2}.
\end{equation}
While the projection defined in Eq.~\eqref{eq:mse_minimizer} is orthogonal with respect to the \( L_2 \) norm, one can generalize this notion using a broader class of loss functions known as Bregman divergence \cite{Rosic2013ParameterSetting, Siahkamari2020LearningDivergence}. 

By solving the previous minimization problem we obtain a mapping \(\varphi\) that assigns to each state \(s\) its best approximation of the value function.

\subsection{Bellman informed conditional expectation}
In practical reinforcement learning problems, we may observe a function \( g(s)\) of the states, rather than the states themselves. In that case, we are interested in estimating
\begin{equation}
    \mathbb{E}[V_\pi(s) \mid \mathcal{B}_g] = \varphi(g(s)),
\end{equation}
where \(\mathcal{B}_g = \sigma(g(s))\) is the \(\sigma\)-algebra generated by the random variable \(g(s)\). A typical example represents the expected collected reward  $r:=\mathbb{E}_\pi(R(s))$. Hence, let \( r := g(s) \) denote the expected reward function evaluated at the state \( s \). Then, following the previous section, one may write:
\begin{equation}
    \mathbb{E}[V_\pi(s) \mid \mathcal{B}_g] = \varphi(r),
\end{equation}
as well as
\begin{equation}
    \label{eqn:gen_filter_r}
    V_\pi(s) = \varphi(r) + \bigl(V_\pi(s) - \varphi(r)\bigr),
\end{equation}
in which the map $\varphi(\cdot)$ can be found by solving following optimization problem:
\beq
\label{opt_la}
\mathbb{E}[V_\pi(s) \mid \mathcal{B}_g]= \arg\min_{\hat{\varphi}} \mathbb{E}\left[\left| V_\pi(s) - \hat{\varphi}(r) \right|^2 \right].
\eeq
We note, however, that here $L_2$ minimization may not always be appropriate. In the Hilbert space framework, the value function \( V_\pi \) is modeled as an element of an $L_2$ space with the corresponding inner product. However, the reward function \( r = g(s) \) may not always satisfy the necessary integrability or boundedness conditions to lie directly in this space. For example, if \( r \) is unbounded or has infinite variance, it may not be a member of $L_2$ space. In such cases one often considers extensions or approximations of the reward function that project \( r \) into a suitable subspace, or employ alternative function spaces (e.g., weighted \( L_p \) spaces) to accommodate unbounded rewards while preserving a meaningful projection framework. This ensures that the projection of \( V_\pi \) onto functions of \( r \) remains well-defined, enabling the estimation of the map \(\varphi\) as the best approximation of \( V_\pi \) measurable with respect to the reward features within the chosen functional setting.

To computationally accommodate the decomposition in Eq.~(\ref{eqn:gen_filter_r}) and the optimization problem in Eq.~(\ref{opt_la}), we need prediction of the reward function.   By using the Bellman equation \cite{Sutton1998}, the value function can be expressed as
\begin{equation}
    \label{eqn:bellman_base}
    V_\pi(s) = r(s) + \gamma \, \mathbb{E}_\pi\left[ V_\pi(S') \mid S = s \right],
\end{equation}
where the expectation over \(S'\) is taken with respect to the transition distribution induced by the policy \(\pi\). The true value function \(V_\pi\) is then the fixed point of the stochastic Bellman operator \(T\):
\[
    V_\pi = T V_\pi,
\]
where
\[
    (T V)(s) := r(s) + \gamma \, \mathbb{E}_\pi \big[ V(S') \mid s \big].
\]
In classical temporal-difference learning \cite{sutton1988td, Bradtke1996}, the Bellman projection is defined as the orthogonal projection of the Bellman operator \(T\) applied to a value function onto the space of representable value functions \(\mathcal{V}_\Theta\):
\[
V_\pi^* = (\Pi T)(V_\pi) = \arg\min_{U \in \mathcal{V}_\Theta} \mathbb{E}(\| U - TV_\pi \|^2)
\]
This ensures that \(V_\pi^\ast\) is the closest function in the space of parametrized value functions to the exact Bellman update, and the fixed point of the projected stochastic Bellman operator satisfies \(\Pi TV_\theta = V_\theta\), where \(\Pi\) is the projection operator \cite{Antos2008, Tziortziotis2017BayesianRegularization}.

In contrast, in our approach, we define a Bellman-informed representation of rewards \(r\) with
\[
r(s) := V_{\pi}(s) - \gamma \mathbb{E}_\pi[V_{\pi}(S') \mid s] 
\]
and search the optimal map in the space of measurable maps:
\begin{equation}
    \label{eqn:mse_projection}
    \varphi^* = \arg\min_{\hat{\varphi}} \mathbb{E}\left[ \| (V_{\pi}(s) - \hat{\varphi}(V_\pi(s) - \gamma \, \mathbb{E}_\pi\left[ V_\pi(S') \mid S = s\right])  \|^2 \right]
\end{equation}
in which the minimizer \(\varphi^*\) is the best approximation under the Bellman target in the MSE sense within the chosen function class. Hence, the projection in Eq.~(\ref{eqn:mse_projection}) is related to the conditional expectations under Bellman condition, and given reward. It is an information-theoretic one, and not a classical projection of Bellman residual, but its inverse problem.
This view allows us to reinterpret classical temporal-difference learning from the perspective of an inverse problem, as further explained in the text.
\section{Gauss-Markov-Kalman filter temporal difference (GMKF-TD)}

Let be given a fixed observation \( \hat{r}(s) \) that provides real measurement information about the true reward \( \tilde{r}(s) \) such that
\beq \hat{r}(s)= \tilde{r}(s)+\hat{\eta}\eeq
holds. Here, \(\hat{\eta}\) is one realization of the measurement noise \(\eta(\omega)\), a zero-mean random variable with \(\mathbb{E}[\eta(\omega)] = 0\), assumed independent of the prior knowledge on $V_{\pi,f}(s,\omega)$ ($f$ here stands for the forecast). Note that $\omega$ is an event in the probability space $(\varOmega,\mathfrak{F},\mathbb{P})$ \footnote{we kept same notation as in the previous section due to simplicity} understood as the product of probability spaces to which belong $V_{\pi,f}(s,\omega)$ and $\eta(\omega)$, respectively. In the classical measurement noise setting, the prior prediction $r_f(s,\omega)$ of the observation \( \tilde{r}(s) \) is modeled as:
\beq
 r_f(s,\omega)= \tilde{r}(s,\omega) + \eta(\omega)
\eeq
in which $\tilde{r}(s,\omega)$ is the prediction of the reward. 

Following the previous section, one may further write the decomposition of the prior knowledge on the value of the state as:
\begin{equation}
    \label{eq:decomposition_final_rep}
    V_{\pi,f}(s, \omega) = \varphi(r_f(s,\omega)) + (V_{\pi,f}(s, \omega) -\varphi(r_f(s,\omega))
\end{equation}
where \(V_{\pi,f}(s, \omega)\) denotes the prior knowledge on the value function, and \(\varphi(r_f(s,\omega))\) represents the best approximation of \( V_{\pi,f} \) given knowledge of the prediction of the reward \(r_f(s,\omega)\). This formulation then leads to:
\beq
V_{\pi,f}(s, \omega) = \varphi(r_f(s,\omega)) + \epsilon(s,\omega),
\eeq
where
\beq
\label{err}
\epsilon(s,\omega) := V_{\pi,f}(s, \omega) - \varphi(r_f(s,\omega))
\eeq
is the residual of our prior knowledge, satisfying
\beq
\mathbb{E}[\epsilon(s,\omega) \mid r_f(s,\omega)] = 0.
\eeq
Our goal is to define an update estimator in Eq.~(\ref{eq:decomposition_final_rep}) that incorporates the real measurement information $\hat{r}(s)$. Conceptually, this amounts to refining the prior knowledge by adjusting the original sample to account for the new observation:
\beq
\label{correction_eq}
 {V}_{\pi,a}(s, \omega) :=\tilde{V}_{\pi,f}(s, \omega \mid \hat{r}) =V_{\pi,f}(s, \omega) +  \mathbb{E}[V_{\pi,f}(s, \omega) \mid \hat{r}]- \mathbb{E}[ V_{\pi,f}(s, \omega) \mid r_f(s,\omega)].
\eeq
Here, ${V}_{\pi,a}(s, \omega)$ denotes the posterior random variable describing the value function after observation of the reward, whereas the term $\mathbb{E}[V_{\pi,f}(s, \omega) \mid \hat{r}]- \mathbb{E}[ V_{\pi,f}(s, \omega) \mid r_f(s,\omega)]$ models correction of the prior information. 
Following Doob-Dynkin lemma one has
\beq
{V}_{\pi,a}(s, \omega) = V_{\pi,f}(s, \omega) + \left( \mathbb{E}[V_{\pi,f}(s, \omega) \mid \hat{r}] - \varphi(r_f(s,\omega)) \right),
\eeq
i.e.
\begin{align}
{V}_{\pi,a}(s, \omega) &= V_{\pi,f}(s, \omega) + \left( \mathbb{E}[\varphi(r_f(s,\omega)) \mid \hat{r}] - \varphi(r_f(s,\omega)) + \mathbb{E}[\epsilon(s,\omega) \mid \hat{r}] \right).
\label{paestyima}
\end{align}
The last equation is obtained by using the linearity of conditional expectation and Eq.~(\ref{err}) giving the following identity:
\beq
\mathbb{E}[V_{\pi,f}(s, \omega) \mid \hat{r}] 
= \mathbb{E}[\varphi(r_f(s,\omega)) \mid \hat{r}] + \mathbb{E}[\epsilon(s,\omega) \mid \hat{r}].
\eeq
\noindent As the residual is mean-zero conditioned on \( \hat{r} \), i.e.,
\[
\mathbb{E}[\epsilon(s,\omega) \mid \hat{r}] = 0,
\] 
and having that  \( \mathbb{E}[\varphi(r_f(s,\omega)) \mid \hat{r}] =\varphi(\hat{r}) \), the estimator in Eq.~(\ref{paestyima}) simplifies to:
\begin{equation}
\label{main_update}
{V}_{\pi,a}(s, \omega) = V_{\pi,f}(s, \omega) + \varphi(\hat{r}) - \varphi(r_f(s,\omega)),
\end{equation}
which is the base formula for further updating. To extract its first two moments, one may take expectation to obtain the mean-value
\begin{equation}
\bar{V}_{\pi,a}(s, \omega) = \mathbb{E}[V_{\pi,a}(s, \omega)|\hat{r}] = \varphi(\hat{r}),
\end{equation}
which ensures that the updated value $V_{\pi,a}(s, \omega)$ reflects the correct posterior mean conditioned on the observed reward. The orthogonal component
\begin{equation}
\label{gmkf}
V_{\pi,f}^\perp(s, \omega) := V_{\pi,f}(s, \omega) - \varphi(r_f(s, \omega))
\end{equation}
is a zero-mean random variable, and represents the remaining prior knowledge, i.e.~uncertainty about the value of the state-value function. This component can be used to define the covariance of the updated value $V_{\pi,a}(s, \omega)$ as:
\begin{equation}
\mathrm{cov}(V_{\pi,a}) = \mathbb{E}[ V_{\pi,f}^\perp \otimes V_{\pi,f}^\perp ] =: C_a,
\end{equation}
as well as the total variance as
\begin{equation}
\mathrm{var}(V_{\pi,a}) = \mathbb{E}[ \| V_{\pi,f}^\perp(\omega) \|^2 ] = \mathrm{tr}(\mathrm{cov}(V_{\pi,a})).
\end{equation}
The updating equation in Eq.~(\ref{gmkf}) is correct for both moments for problems described by linear mapping and Gaussian distribution. However, in a non-Gaussian case as well as in nonlinear problems, the posterior does not have the correct second order moment \cite{Matthies2016}. Therefore, we need to further generalize the previous formulation. Let be given the random variable
 \[W_{\pi,f}:=\Psi(V_{\pi,f}(s, \omega))\] defined by the continuous mapping $\Psi: V_{\pi,f} \to W_{\pi,f}$ of the random variable $V_{\pi,f}(s, \omega)$ to $W_{\pi,f}$. Analogous to the previous theory, the conditional expectation of this random variable can be expressed as:
\begin{equation}
\label{general_map}
\mathbb{E}\big[\Psi(V_{\pi,f}(s, \omega)) \mid \sigma(r_f(s_{t+1},\omega))\big] = \varphi_\Psi(r_f(s_{t+1},\omega))
\end{equation}
in which \(\varphi_\Psi(\cdot)\) is the optimal map with respect to the transformation $\Psi(\cdot)$.
Once the actual reward observation $\hat{r}$ becomes available, the posterior knowledge of $W_{\pi,f}(s, \omega)$, $W_{\pi,a}(s, \omega)$, can be updated as
\begin{equation}
\label{main_update_higher}
W_{\pi,a}(s, \omega) = W_{\pi,f}(s, \omega) + \varphi_\Psi(\hat{r}) - \varphi_\Psi(r_f(s,\omega)),
\end{equation}
with the posterior mean given as:
\begin{equation}
\mathbb{E}\big[W_{\pi,a}(s,\omega) \mid \hat{r} \big] = \varphi_\Psi(\hat{r}).
\end{equation}
The last equation then reduces to the previously described update in Eq.~(\ref{main_update}) by taking 
\(W_{\pi,f}= V_{\pi,f}(s, \omega)\) in which the map \(\Psi(V_{\pi,f}(s, \omega)) \equiv V_{\pi,f}(s, \omega)\), $\Psi(\cdot)=\textrm{Id}(\cdot)$ and $\varphi(\cdot)$ is the same as $\varphi_{Id}$. 
If we further define
\[
W_{\pi,f} := \Psi(V_{\pi,f}(s,\omega)) = \big\| V_{\pi,f}(s,\omega) - \varphi(r_f(s,\omega)) \big\|^2,
\]
then Eq.~(\ref{main_update_higher}) yields the total conditional variance
\beq
\mathrm{Var}(V_{\pi,a} \mid \hat{r}) 
= \mathbb{E}\Big[ \big\| V_{\pi,f}(s,\omega) - \varphi(r_f(s,\omega)) \big\|^2 \,\big|\, \hat{r} \Big]
= \varphi_{\mathrm{VAR}}(\hat{r}) - \big( \bar{V}_{\pi,a}(\hat{r}) \big)^2,
\eeq
where 
\beq
\mathbb{E}\big[ V_{\pi,f}^2 \mid \hat{r} \big]=\varphi_{\mathrm{VAR}}(\hat{r}) , 
\eeq
represent the optimal conditional second moment represented by a mapping $\varphi_{\mathrm{VAR}}(\cdot)$. Therefore, to ensure that the update in Eq.~(\ref{gmkf}) yields the correct posterior mean and variance, the equation must be modified to:
\begin{equation}
\label{corr_var}
V_{\pi,a}^I = \varphi(\hat{r}) + \frac{\mathrm{var}(V_{\pi,a})}{\mathrm{VAR}(V_{\pi,a})} 
\big( V_{\pi,f}(s,\omega) - \varphi(r_f(s,\omega)) \big).
\end{equation}
Following this, we may further design an update equation for the correct mean and the covariance \cite{Matthies2016} as:
\begin{equation}
\label{main_gmkf}
V_{\pi,a}^{II}(s,\omega) = \varphi(\hat{r}) + B_V B_1^{-1} \big(V_{\pi,f}(s,\omega) - \varphi(r_f(s,\omega))\big).
\end{equation}
Here, we design the correct posterior covariance by taking 
\beq
W_{\pi,f}:=\Psi(V_{\pi,f}(s,\omega)) := \big(V_{\pi,f}(s,\omega) - \varphi(r_f(s,\omega))\big)^{\otimes 2}
\eeq
which allows us to compute the posterior covariance of the updated value function:
\beq
\label{eq_c_v}
\operatorname{cov}(V_{\pi,a}) = \mathbb{E}\Big[\big(V_{\pi,f}(s,\omega) - \varphi(r_f(s,\omega))\big)^{\otimes 2} \,\big|\, \hat{r} \Big] = \psi_{\otimes 2}(\hat{r}) =: C_V
\eeq
given optimal map $\psi_{\otimes 2}(\cdot) $. 
In Eq.~(\ref{main_gmkf}) $B_1$ denotes the “square root” satisfying 
\[B_1 B_1^* = C_{a},\]
and similarly 
\begin{equation}
\label{square_root}
B_V B_V^* = C_V.
\end{equation} 
The square root can be computed by Cholesky or eigen-value decomposition \cite{Matthies2016}. 

The transformation in Eq.~(\ref{main_gmkf}) is linear, and only preserves the first two moments. One can further extend the equation to update the skewness as well as kurtosis, but that would require definition of higher order tensors, which could be computationally expensive. In this paper we limit ourselves on the first two moments as these are used in most of practical engineering applications.

In the previous equations, we have derived estimation of the first and the second moment of the value function. However, these equations are not yet designed for computational practice. Both of update equations require the knowledge of the future reward, that has to be predicted. 
Using the Bellman equation one may further express the future reward:
\begin{equation}
\label{main_update_BellmanA}
r_f(s,\omega):= \left({V}_{\pi,f}(s, \omega) - \gamma \mathbb{E}_\pi[V_{\pi,f}(S',\omega)|S=s]\right)+\eta(\omega)  \end{equation}
leading to the update equation for the mean:
\begin{equation}
\label{main_update_Bellman}
{V}_{\pi,a}(s, \omega) =V_{\pi,f}(s, \omega)+\varphi(\hat{r}) - \varphi(r_f(s, \omega)),
\end{equation}
in which the map $\varphi(\cdot)$ is now the optimal map under Bellman target, see Eq.~(\ref{eqn:mse_projection}). Similarly, one may reformulate the update equation for the covariance 
\begin{equation}
\label{main_gmkf_bel}
V_{\pi,a}^{II}(s,\omega) = \varphi(\hat{r}) + B_V^b B_1^{-1} \big(V_{\pi,f}(s,\omega) - \varphi(r_f(s,\omega))\big),
\end{equation}
in which $B_V^b$ is the covariance obtained by optimal map under Bellman target, see Eq.~(\ref{general_map}) and Eq.~(\ref{square_root}).

\subsection{Optimal map for the mean}
For computational purposes in Eq.~(\ref{main_update_Bellman}) one has to assume family of measurable functions $\varphi(\cdot)$. A common and tractable choice is a linear (affine) form, $\varphi(r)= Kr + b$, where $K$ corresponds to the optimal Kalman gain and $b$ is a bias term \cite{Matthies2016}. The coefficients of a linear map then can be computed by minimizing the orthogonal residual \cite{Luenberger1968, Bojana2012}
\begin{eqnarray}
    (K, b) ={\arg \min}_{K,b} \|V_{\pi,f}(s,\omega) - (Kr_f(s,\omega)+b)\|^2
\end{eqnarray}
in which the reward prediction $r_f(s,\omega)$ is given in Eq.~(\ref{main_update_BellmanA}) such that:
\begin{equation}
\label{main_update_Bellman_kf}
{V}_{\pi,a}(s, \omega) =V_{\pi,f}(s,\omega)+K(\hat{r} -{V}_{\pi,f}(s, \omega) + \gamma \mathbb{E}_\pi[V_{\pi,f}(S',\omega)|S=s]+\eta(\omega)),
\end{equation}
holds. The optimal gain is defined as
\begin{eqnarray}
    \label{eqn:kalman_gain}
    K = C_{V_{\pi,f}, r_f(s,\omega)}(C_{r_f(s,\omega)})^\dagger
\end{eqnarray}
in which $C_{V_{\pi,f}, r_f(s,\omega)}$ is the covariance between the prior on the state value function and the forecasted reward, whereas $C_{r_f(s,\omega)}$ is the covariance of the forecasted reward.
The above filter is called the Gauss-Markov Kalman filter based Temporal Difference (GMKF-TD) as it is an extract of the GMKF theory \cite{Matthies2016}, implemented for temporal difference. By taking the expectation of the previous equation:
\begin{equation}
\label{main_update_Bellman_kf1}
{V}_{\pi}(s) =V_{\pi}(s)+K(\hat{r} -{V}_{\pi}(s) + \gamma \mathbb{E}_\pi[V_{\pi}(S',\omega)|S=s]))
\end{equation}
in which $V_{\pi}(s):=\mathbb{E}(V_{\pi,f}(s,\omega))$,
one obtains the equation that resembles the temporal difference algorithm
\begin{equation}
\label{main_td}
{V}_{\pi}(s) =V_{\pi}(s)+\alpha(\hat{r} -{V}_{\pi}(s) + \gamma \mathbb{E}_\pi[V_{\pi}(S',\omega)|S=s])),
                                     \end{equation}
if the Kalman gain $\alpha$ is to be interpreted as learning rate. Hence, the innovation term \( \hat{r} - r_f(s, \omega) \) in Eq.~(\ref{main_update_Bellman_kf1}) can be viewed as a generalized expression of the temporal difference (TD) error, where the Kalman gain $K$ plays a role of automatic choosing of the learning rate. Note that only in a linear case these two algorithms match. Otherwise, in a nonlinear case (by choosing the optimal map as a neural network or polynomial) our algorithm generalizes the temporal difference for nonlinear examples.

To implement the GMKF-TD algorithm, we need to deal with the expectation term for future returns in Eq.~(\ref{main_update_Bellman_kf}). In practice one usually approximates it with one sample, i.e.~the full distribution of states $ S'$ is substituted by one particular one:
\[
\mathbb{E}_\pi[V_{\pi,f}(S', \omega) \mid S = s] 
\approx V_{\pi,f}(s', {\omega}),
\]
with an implicit zero-mean noise term \cite{Engel2005, Geist2010}:
\[
\varepsilon_{V'} = V_{\pi,f}(s', \omega)
- \mathbb{E}_\pi[V_{\pi,f}(S', \omega) \mid S = s]
\]                                                          
such that
\[
V_{\pi,f}(s', {\omega})=\mathbb{E}_\pi[V_{\pi,f}(S', \omega) \mid S = s]+\varepsilon_{V'} 
\]
holds.
One may assume that this error follows normal distribution with constant variance for the stationary environment.

Following previous discussion, one may further write:
\begin{equation}
\label{main_update_Bellman_kf_final}
{V}_{\pi,a}(s, \omega) =V_{\pi,f}(s,\omega)+K(\hat{r} -{V}_{\pi,f}(s, \omega) + \gamma V_{\pi,f}(s',\omega)|S=s]+\varepsilon_r)
\end{equation}
in which $\varepsilon_r= \gamma \varepsilon_V+\eta$ consists of both modelling and observation errors. 

Note that the linear mapping as described previously is not optimal in nonlinear cases. In such a case one may choose neural network approximation or other nonlinear mappings for $\varphi(\cdot)$ \cite{Matthies2016}. 

\subsection{Update equation for the covariance}

Similarly to the update for the mean, Eq.~(\ref{main_gmkf}) is not practically useful due to definition of abstract optimal measurable mappings. In terms of covariance, we choose mapping $\Psi(\cdot)$ that is positive definite following covariance properties. There are many ways of achieving this, but in this paper we choose simple mapping representing the lower triangular decomposition:
\beq
\label{eq:low_triang}
\Psi(r_f) = L(r_f) L(r_f)^\top, \quad L(r_f) = M_0 + r_f M_1,
\eeq
where $M_0, M_1 \in \mathbb{R}^{d \times m}$ such that $\Psi(r_f) \succeq 0$ automatically. The coefficients of the mapping then can be computed by solving the following minimization problem expressed in terms of Frobenius norm:
\beq
\label{eq:cov_min}
\min_{M_0, M_1} \mathbb{E} \Big[ \big\| \tilde{V}_{\pi,f} \otimes \tilde{V}_{\pi,f} - (M_0 + r_f M_1)(M_0 + r_f M_1)^\top \big\|_F^2 \Big].
\eeq
This optimization problem is nonlinear in parameters, but guarantees positive semi-definiteness for all $r$ \cite{Ju2026SPDChallenges, Vishny2024High-DimensionalSamples}. Note that the mapping can be also extended to neural network approximation that guarantees positive semi-definiteness \cite{Parish2024EmbeddedFasteners}.  
Specifically, let $L_w(r)$ be a neural network with parameters $w$ that outputs the lower triangular matrix, and define
\beq
\Psi_\theta(r_f) = L_w(r_f) L_w(r_f)^\top.
\eeq
By construction, $\Psi_\theta(r_f) \succeq 0$ for all $r_f$, and the parameters $w$ can be learned by minimizing the Frobenius-norm loss
\[
\min_\theta \mathbb{E}\Big[ \big\| \tilde{V}_{\pi,f} \otimes \tilde{V}_{\pi,f} - \Psi_\theta(r_f) \big\|_F^2 \Big].
\]
This approach allows for flexible, nonlinear dependence of the covariance on the reward $r_f$ while ensuring positive semi-definiteness.

Once the map is built, one can perform the update
\begin{equation}
\label{main_gmkf_cov}
V_{\pi,a}^{II}(s,\omega) = \varphi(\hat{r}) + B_V B_1^{-1} \big(V_{\pi,f}(s,\omega) - \varphi(r_f(s,\omega))\big).
\end{equation}
which for the linear mapping reads:
\begin{equation}
\label{main_gmkf_cov_lin}
V_{\pi,a}^{II}(s,\omega) = K\hat{r} + B_V B_1^{-1} \big(V_{\pi,f}(s,\omega) - Kr_f(s,\omega)\big).
\end{equation}
The square root covariance $B_1^{-1}$ is the one computed given covariance of Eq.~(\ref{main_update_Bellman_kf_final}), whereas the square-root covariance $ B_V$ comes given the mapping $\Psi_\theta(r_f)$ and can be computed using Eq.~(\ref{eq_c_v}).

\section{Sequential learning for non-stationary environments}

If the environment is non-stationary (i.e.~transitions or rewards change over time), then the state-value function is known to be time-dependent.
Let $V_{\pi,t}(s)$ denote the value of state $s$ at time $t$ defined as
\begin{equation}
V_{\pi,t}(s) = \mathbb{E}_\pi \left[ \sum_{k=0}^{\infty} \gamma^k \, R_{t+k+1} \,\middle|\, S_t = s \right],
\end{equation}
where the rewards $R_{t+k+1}$ and the transition probabilities $P_t(s_{t+k+1} | s_{t+k})$ are explicitly time-dependent. The Bellman equation then reads
\beq
V_{\pi,t}(s) = \mathbb{E}_\pi \Big[ R_{t+1} + \gamma V_{\pi,t+1}(S_{t+1}) \,\big|\, S_t = s \Big].
\eeq

To model the unknown value function \(V_{\pi,t}(s)\) probabilistically, we treat $V_{\pi,f}$ a priori as a continuous-time stochastic process on a probability space $(\varOmega, \mathcal{F}, \mathbb{P})$, indexed by time $t \in \mathcal{T} = [0,T]$ and parameterized by state $s$:
\[
V_{\pi,f}(t,s,\omega) \in \mathcal{V}, \quad \omega \in \varOmega,
\]
where $\varOmega$ is the sample space, $\mathcal{F}$ is a sigma-algebra of events, $\mathbb{P}$ is the probability measure, and $\mathcal{V}$ is the space of values of the state-value function. For each fixed $t \in \mathcal{T}$, we define
\beq
V_{\pi,f,t}(s,\omega) := V_{\pi,f}(t,s,\omega) \in L_2(\varOmega, \mathcal{F}_t, \mathbb{P}),
\eeq
where $\{\mathcal{F}_t\}_{t \in \mathcal{T}}$ is a filtration $
\mathcal{F}_0 \subseteq \mathcal{F}_{t_1} \subseteq \dots \subseteq \mathcal{F}_T \subseteq \mathcal{F}.$ Then, the stochastic process can be seen as the family of random variables
$\{ V_{\pi,f,t}(s,\omega) \}_{t \in \mathcal{T}}$
defined on $(\varOmega, \mathcal{F}, \mathbb{P})$ and adapted to the filtration $\{\mathcal{F}_t\}$. If we discretize time into $\{t, t+1, \dots, t+n\} \subset \mathcal{T}$, where $t+k$ denotes the $k$-th discrete time step after $t$, the discrete-time stochastic process is
\[
\{ V_{\pi,f,t+k}(s,\omega) \}_{k=0}^{n}, \quad V_{\pi,f,t+k}(s,\omega) := V_{\pi,f}(t+k,s,\omega),
\]
with each $V_{\pi,f,t+k}(s,\omega)$ being $\mathcal{F}_{t+k}$-measurable.

After assuming the prior knowledge on $V_{\pi,t}$, the temporal difference update rule in Eq.~(\ref{main_update_Bellman_kf}) then can be extended to the non-stationary environment as
\begin{equation}
\label{main_update_Bellman_kf_time}
{V}_{\pi,a,t+1}(s, \omega) =V_{\pi,f,t}(s,\omega)+K(\hat{r}_{t+1} -{V}_{\pi,f,t}(s, \omega) + \gamma \mathbb{E}_\pi[V_{\pi,f,t+1}(S_{t+1},\omega)|S_t=s]+\eta_t(\omega))
\end{equation}
in which $\eta_t(\omega)$ represents the observation noise and is independent of the prior information. Here, the probability space is to be understood as the product space $(\varOmega', \mathcal{F}', \mathbb{P}') := 
\big(\varOmega \times \varOmega_\eta  \; 
\mathcal{F} \otimes \mathcal{F}_\eta, \;
\mathbb{P} \otimes  \mathbb{P}_\eta \big)
$ with realizations denoted by $\omega$. The previous equation has sequential form in which the next prior is obtained from the previous posterior e.g.~by random walk:
\beq
{V}_{\pi,f,t+1}(s,\omega) = V_{\pi,a,t}(s,\omega) + w_t(\omega), \quad w_t(\omega) \perp \mathcal{F}_t, \quad \mathbb{E}[w_t(\omega)] = 0,
\eeq
in which $w_t(\omega)$ is independent random variable from ${V}_{\pi,a,t}(s, \omega)$.
In practice, the expectation 
\(
\mathbb{E}_\pi \big[ V_{\pi,f,t}(S_{t+1}, \omega) \,\big|\, S_t = s \big]
\) in Eq.~(\ref{main_update_Bellman_kf_time})
is generally intractable, so it is replaced by a single sampled next-state value 
\(
V_{\pi,f,t+1}(s', \omega),
\)
leading to the sampled TD update:
\begin{equation}
\label{main_update_Bellman_kf_time_final}
{V}_{\pi,a,t+1}(s, \omega) =V_{\pi,f,t}(s,\omega)+K(\hat{r}_{t+1} -{V}_{\pi,f,t}(s, \omega) + \gamma V_{\pi,f,t+1}(s',\omega) +\varepsilon_{r,t}(\omega)).
\end{equation}
Here, $\varepsilon_{r,t}(\omega):=\varepsilon_{V}^t (\omega)+\eta(\omega)$ represents both the modelling and measurement error.
The modelling error is the discrepancy between the sampled version of the conditional expectation and its true value, i.e.
\beq
\label{eqn:epsilon_V}
\varepsilon_{V}^t(\omega) = V_{\pi,f,t}(s', \omega)
- \mathbb{E}_\pi[V_{\pi,f,t}(S_{t+1}, \omega) \mid S_t = s]
\eeq
Here, the probability space is to be understood as the product space $(\varOmega', \mathcal{F}', \mathbb{P}') := 
\big(\varOmega \times \varOmega_\eta \times \varOmega_\xi,  \; 
\mathcal{F} \otimes \mathcal{F}_\eta \otimes \mathcal{F}_\xi, \;
\mathbb{P} \otimes  \mathbb{P}_\eta  \otimes  \mathbb{P}_\xi \big)
$ with realizations denoted by $\omega$.
To capture temporal dependence in the modeling error, we consider the stochastic process $\{\varepsilon_V^t\}_{t \geq 0}$ and decompose it into a predictable component and an innovation term. Specifically, for each $t$, we write
\begin{equation}
\varepsilon_V^{t+1}
= \mathbb{E}\!\left[\varepsilon_V^{t+1} \mid \mathcal{F}_t^{'}\right]
+ \xi_t,
\quad
\xi_t := \varepsilon_V^{t+1} - \mathbb{E}\!\left[\varepsilon_V^{t+1} \mid \mathcal{F}_t^{'}\right],
\end{equation}
where $\{\mathcal{F}_t^{'}\}$ denotes the underlying filtration. By construction, the innovation process $\xi_t$ satisfies $\mathbb{E}[\xi_t \mid \mathcal{F}_t^{'}] = 0$, and is therefore a martingale difference sequence representing the component of $\varepsilon_V^{t+1}$ that is not predictable from past information. To obtain a tractable model, we approximate the conditional expectation by a linear function of the current error, i.e.,
\begin{equation}
\mathbb{E}\!\left[\varepsilon_V^{t+1} \mid \mathcal{F}_t^{'}\right] \approx \rho \, \varepsilon_V^t,
\end{equation}
which yields the autoregressive AR(1) model \cite{Antos2008, Engel2005}
\begin{equation}
\varepsilon_V^{t+1} = \rho \, \varepsilon_V^t + \xi_t,
\end{equation}
with $\rho \in [0,1]$ controlling the temporal correlation and $\xi_t$ capturing the unpredictable innovation at each time step.

Similarly to the update equation for the mean, we may also update the covariance following same derivation as for the mean. The update equation then reads:
\begin{equation}
\label{main_gmkf_bel-time}
V_{\pi,a,t}^{II}(s,\omega) = \varphi(\hat{r}) + B_{V,t}^b B_{1,t}^{-1} \big(V_{\pi,f,t}(s,\omega) - \varphi(r_f(s,t,\omega))\big),
\end{equation}
in which $\varphi(\cdot)$ is the optimal map for the conditional expectation (here linear), and $B_{V,t}^b$, $B_{1,t}$ are the square roots of covariances obtained by optimal maps under Bellman target (see Section 5.2), Eq.~(\ref{general_map}) and Eq.~(\ref{square_root}).
\section{Parametrization and discretization of GMKF-TD}

In a stationary environment, we model prior knowledge of the value function using a linear parameterization of the form
\begin{equation}
  \label{eqn:lin_v}
    V_\pi(s) = \bm{\theta}^\top \bm{\Phi}(s),
\end{equation}
where $\bm{\theta} \in \mathbb{R}^M$ are the parameters and $\bm{\Phi}(s) = [\Phi_1(s), \dots, \Phi_M(s)]^\top \in \mathbb{R}^M$ is a vector of measurable feature functions, commonly radial basis functions, polynomial expansions, and Fourier series \cite{Amherst2008ValueBasis}. To capture uncertainty in the value function, we consider $V_{\pi,f}$ as a stochastic variable on the probability space $(\varOmega, \mathcal{F}, \mathbb{P})$, and we introduce the random parameter vector with finite second-order moments $\bm{\theta}_{f,t} : \varOmega \to \mathbb{R}^M$ such that
\begin{equation}
  \label{eqn:v_param}
    V_{\pi,f}(s,\omega) := (\bm{\theta}_{f}(\omega))^\top \bm{\Phi}(s), \;\; \forall \omega \in \varOmega
\end{equation}
holds. To account for non-stationary dynamics in the environment, we model the time evolution of the parameters as a random walk:
\begin{equation}
 \label{eqn:theta_t}
    \bm{\theta}_{f,t}(\omega) = \bm{\theta}_{f,t-1}(\omega) + \bm{\nu}_{f,t}(\omega), \quad t \ge 1.
\end{equation}
 The increments $\{\bm{\nu}_{f,t}\}_{t \ge 1}$ are independent random variables on a joint probability space $(\varOmega, \mathcal{F}, \mathbb{P})$ \footnote{Here we use same notation but space is defined for stochastic process not variable}, each with zero mean $\mathbb{E}[\bm{\nu}_{f,t}] = 0$ and finite variance $\mathbb{E}[\|\bm{\nu}_{f,t}\|^2] < \infty$. This then leads to representation of the prior state value function:
 \begin{equation}
  \label{eqn:lin_v1}
    V_{\pi,f,t}(s) = \bm{\theta}_{f,t}^\top(\omega) \bm{\Phi}(s).
\end{equation}
Substituting the previous parameterization into Eq.~(\ref{main_update_Bellman_kf_time_final}), one obtains
 \begin{equation}
\label{eq:td_update_param}
\bm{\theta}_{a,t+1}^\top(\omega) \bm{\Phi}(s)
=
\bm{\theta}_{f,t}^\top(\omega)  \bm{\Phi}(s)
+ K_t \Big(
\hat r_{t+1} - \bm{\theta}_{f,t}^\top(\omega)  \bm{\Phi}(s) 
+ \gamma \, \bm{\theta}_{f,t+1}^\top(\omega)  \bm{\Phi}(s') 
- \varepsilon_{r,t}
\Big).
\end{equation}
Note that
\begin{eqnarray}
\label{eq:td_operator}
r_{f,t+1}(\omega)
&=&\bm{\theta}_{f,t}^\top(\omega)  \bm{\Phi}(s) - \gamma \, \bm{\theta}_{f,t+1}^\top(\omega)  \bm{\Phi}(s')  + \varepsilon_{r,t}(\omega) \nonumber \\
&=&\bm{\theta}_{f,t}^\top(\omega)  \bm{\Phi}(s) - \gamma \, (\bm{\theta}_{f,t}(\omega)+\bm{\nu}_{f,t+1}(\omega))^\top  \bm{\Phi}(s')  + \varepsilon_{r,t}(\omega)
\end{eqnarray}
which can be grouped as
\begin{equation}
\label{eq:td_operator1}
H_t^T \bm{\theta}_{f,t}(\omega)=(\bm{\Phi}(s) - \gamma \, \bm{\Phi}(s'))^T\bm{\theta}_{f,t}(\omega),
\end{equation}
as well as
\begin{equation}
\label{eq:td_error1}
\epsilon_{r,t}(\omega)=\bm{\nu}_{f,t+1}(\omega)^\top \gamma  \bm{\Phi}(s')  + \varepsilon_{r,t}(\omega).
\end{equation}
Then, the Kalman gain reads:
\begin{equation}
\label{eq:scalar_K}
K_t =
\bm{\Phi}(s)^\top \mathrm{Cov}(\bm{\theta}_{f,t}) \, H_t \,
\left(
H_t^\top \mathrm{Cov}(\bm{\theta}_{f,t}) \, H_t + R_t
\right)^{-1}.
\end{equation}
where $R_t = \mathrm{Cov}(\epsilon_{r,t})$ is the  noise covariance that further can be written as:
\begin{equation}
\label{eq:noise_cov}
R_t =
\gamma^2 \bm{\Phi}(s')^\top Q_{t+1} \, \bm{\Phi}(s') + \mathrm{Cov}(\varepsilon_{r,t+1}(\omega))
\end{equation}
and $Q_{t+1} = \mathrm{Cov}(\bm{\nu}_{f,t+1})$ is the prior covariance capturing non-stationarity.  Intuitively, $K_t$ scales the TD error according to the total uncertainty in the target: it increases the update when the parameters are uncertain or the reward is reliable, and decreases it when the parameters are confident or the reward is noisy. The inclusion of the term $ Q_{t+1} $ ensures that uncertainty in the future value estimate is incorporated, which is critical in non-stationary environments.

The TD update can also be written directly in terms of the parameter vector, without explicitly including the basis functions in the update equation, by absorbing the features into a suitable gain matrix \(K_t\):
\begin{equation}
\label{eq:theta_direct_update_text}
\bm{\theta}_{a,t+1} 
= \bm{\theta}_{f,t} 
+ \hat{K}_t \Big( \hat r_{t+1} + \gamma \, \bm{\theta}_{f,t+1}^\top \bm{\Phi}(s') - \bm{\theta}_{f,t}^\top \bm{\Phi}(s) - \varepsilon_{r,t} \Big).
\end{equation}
Here, \(\hat{K}_t \) is a gain matrix that incorporates the influence of the feature vector \(\bm{\Phi}(s)\) and the uncertainty in \(\bm{\theta}_{f,t}\):
\begin{equation}
\label{eq:kalman_gain}
\hat{K}_t =\mathrm{Cov}(\bm{\theta}_{f,t}) \, H_t \,
\left(
H_t^\top \mathrm{Cov}(\bm{\theta}_{f,t}) \, H_t + R_t
\right)^{-1}.
\end{equation}
 It is important to note that the two formulations of the TD update are not strictly equivalent in general. While Eq.~\eqref{eq:td_update_param} expresses the update in terms of the scalar TD error projected onto the value function for a given state, Eq.~\eqref{eq:theta_direct_update_text} updates the parameter vector $\bm{\theta}$ directly using a gain matrix $\hat{K}_t$ that absorbs the feature vector. This equivalence holds exactly only if the feature basis $\bm{\Phi}(s)$ is orthogonal, so that each component of $\bm{\theta}$ is affected independently by the TD error. For non-orthogonal features, the TD error associated with one state influences multiple components of $\bm{\theta}$ due to correlations between basis functions, and a naive choice of 
\begin{equation}
\hat{K}_t := \bm{\Phi}(s)K_t,
\end{equation}
does not preserve the value update exactly. In such cases, a matrix gain that accounts for the feature covariance is required to ensure that the parameter update correctly corresponds to the underlying value function update. 

In a stationary environment, the parameters are assumed constant over time, i.e., there is no process noise:
\[
\bm{\theta}_{f,t+1} = \bm{\theta}_{f,t}, \qquad Q_{t+1} = \mathbf{0}.
\]
The temporal-difference (TD) update of the parameter vector $\bm{\theta}_{a,t+1}$ then becomes
\begin{equation}
\label{eq:stationary_td_update}
\bm{\theta}_{a,t+1} = \bm{\theta}_{f,t} + K_t \Big( \hat r_{t+1} - \bm{\theta}_{f,t}^\top \bm{\Phi}(s) + \gamma \, \bm{\theta}_{f,t}^\top \bm{\Phi}(s') - \varepsilon_{r,t} \Big),
\end{equation}
where the Kalman gain vector is defined as
\begin{equation}
\label{eq:stationary_kalman_gain}
\hat{K}_t =\mathrm{Cov}(\bm{\theta}_{f,t}) \, H_t \,
\left(
H_t^\top \mathrm{Cov}(\bm{\theta}_{f,t}) \, H_t + R_t
\right)^{-1}
\end{equation}
in which $R_t = \mathrm{Cov}(\varepsilon_{r,t})$ is the reward noise covariance. The next-state features $\bm{\Phi}(s')$ still appear in the TD target, but there is no additional contribution from future parameter uncertainty, since the environment is stationary. In sequential type of update, the previous posterior $\bm{\theta}_{a,t}$ becomes the new prior plus e.g. independent random variable $\nu_{f,t+1}$ for exploration.

The previously described value function in Eq.~(\ref{eqn:lin_v}) or Eq.~(\ref{eqn:lin_v1}) is continuous in the stochastic space. Thus, a discretization is required. 
A straightforward discretization approach is to use sampling methods, such as basic Monte Carlo \cite{Evensen2019}. 
An ensemble $\Theta = [\theta(\omega_1), \dots, \theta(\omega_J)]$ is then constructed to represent a finite, discretized distribution of the parameter. Given Eq.~(\ref{main_update_Bellman_kf_final}), each sample can be updated directly using the linear filter equation:
\begin{equation}
\label{eq:td_update_paramSAMp}
\bm{\theta}_{a,t+1}^\top(\omega_i) \bm{\Phi}(s)
=
\bm{\theta}_{f,t}^\top(\omega_i)  \bm{\Phi}(s)
+ K_t^s \Big(
\hat r_{t+1} - r_{f,t+1}(\omega_i),
\Big), \quad i=1,..,J.
\end{equation}
The Kalman gain $K_t^s$ \footnote{The upper index 's' denotes the sampling version of the Kalman gain} can be estimated statistically by computing the corresponding covariances given the finite ensemble of samples \cite{Rosic2013ParameterSetting}.

Although practical, the ensemble based GMKF-TD is prone to statistical errros. To circumvent this, one may use functional approximation instead. In this paper we choose the polynomial chaos expansion (PCE) \cite{Xiu2002} as a discretization approach such that
\begin{equation}
    \label{eqn:pce}
    \bm{\theta}_{f,t}(\omega) = \sum_{\alpha \in \mathcal{J}} \bm{\theta}_{f,t}^{(\alpha)} {\varPsi}_\alpha(\bm{\xi}(\omega))
\end{equation}
holds. Here, $\bm{\theta}_{f,t}^{(\alpha)}$ are the unknown deterministic coefficients, $\bm{\xi}(\omega)$ is a vector of known standard random variables, ${\varPsi}_\alpha(\cdot)$ is the orthogonal basis function orthogonal with respect to the probability measure of $\bm{\xi}$, and $\alpha$ is the set of multi-incides \cite{Bojana2012}. Hence, the value function can be further discretized as:
$${V}_{\pi,f,t}(s,\omega) =  \sum_{\alpha \in \mathcal{J}} (\bm{\theta}_{f,t}^{(\alpha)}{\varPsi}_\alpha(\bm{\xi}(\omega)))^T\bm{\Phi}(s) $$
Furthermore, one may introduce 
\[
\tilde{\theta}_{f,t}^{(\alpha)}(s) := (\bm{\theta}_f^{(\alpha)})^\top \, \bm{\Phi}(s)
\]
such that
\[
V_{\pi,f,t}(s,\omega) = \sum_{\alpha \in \mathcal{J}} \tilde{\theta}_{f,t}^{(\alpha)}(s) \, {\varPsi}_\alpha(\bm{\xi}(\omega))
\]
holds. Then, the update equation for the value function parameters in Eq.~(\ref{eq:theta_direct_update_text}) reads:
\begin{equation}
\label{main_update_Bellman_kf_final_discr}
 \tilde{\theta}_a^{(\alpha)}(s)=\tilde{\theta}_f^{(\alpha)}(s)+K(\hat{r}^{(\alpha)} -\tilde{\theta}_f^{(\alpha)}(s) + \gamma \tilde{\theta}_f^{(\alpha)}(s')-\epsilon_{r,t}^{(\alpha)}).
\end{equation}
Gathering all coefficients 
\[
\bm{\tilde{\theta}}_f(s) :=
\begin{bmatrix}
\tilde{\theta}_f^{(\alpha_1)}(s) \\
\tilde{\theta}_f^{(\alpha_2)}(s) \\
\vdots \\
\tilde{\theta}_f^{(\alpha_{|\mathcal{J}|})}(s)
\end{bmatrix} \in \mathbb{R}^{|\mathcal{J}|}, \quad
\bm{\tilde{\theta}}_a(s) :=
\begin{bmatrix}
\tilde{\theta}_a^{(\alpha_1)}(s) \\
\tilde{\theta}_a^{(\alpha_2)}(s) \\
\vdots \\
\tilde{\theta}_a^{(\alpha_{|\mathcal{J}|})}(s)
\end{bmatrix} \in \mathbb{R}^{|\mathcal{J}|}
\]
\[
\bm{\epsilon}_{r} :=
\begin{bmatrix}
\epsilon_{r,t}^{(\alpha_1)} \\
\epsilon_{r,t}^{(\alpha_2)} \\
\vdots \\
\epsilon_{r,t}^{(\alpha_{|\mathcal{J}|})}
\end{bmatrix} \in \mathbb{R}^{|\mathcal{J}|}.
\]
one may further write:
\[
\tilde{\bm{\theta}}_a (s)= \tilde{\bm{\theta}}_f(s) + K_t^{(\alpha)} \, 
\Big( \bm{\hat{r}}  - \tilde{\bm{\theta}}_f (s)
+ \gamma \, \tilde{\bm{\theta}}_f(s') +\bm{\epsilon}_r \Big).
\]
The Kalman gain $K_t^{(\alpha)}$ then can be computed directly given the corresponding covariance matrices (see Eq.~(\ref{eqn:kalman_gain})) by the PCE coefficients, using simple algebraic formulas as shown in \cite{Rosic2013ParameterSetting}. Note that $\bm{\hat{r}} $ is also represented here as the vector of PCE coefficients; however, this vector contains only the first coefficient, which corresponds to the mean value, while all other coefficients are zero since the observation is deterministic.

In the previous update equation, one must account for the noise term $\epsilon_{r,t}$. This term can cause the growth of the PCE basis if each update step introduces additional independent random variables. To address this issue, one can use the square-root formulation of the assimilated parameters as shown in \cite{Pajonk2013}, in which the PCE basis cardinality is kept constant, but the coefficients are linearly transformed to allow for the correction of moments given the additional covariance terms in the noise term.

\section{Action value function and exploration}

While the state-value function $V_\pi(s)$ provides a compact characterization of the expected return under policy $\pi$, it does not fully capture the policy-induced decision structure, as it does not explicitly account for the selected action. In contrast, the action-value function $Q_\pi(s,a)$ yields a more expressive representation by conditioning on both the current state and the executed action:
\begin{equation}
    Q_{\pi,t} (s,a) = \mathbb{E}_\pi \big[ G_t \mid S_t = s, A_t = a \big].
\end{equation}
Therefore, one requires GMKF-TD formulation for the action-value function. The derivation of the filter for the action-value function proceeds analogously to the state-value case and follows the same sequence of steps. For brevity, these derivations are not repeated here. Under the assumption of an affine observation model, an update equation analogous to Eq.~(\ref{main_update_Bellman_kf_final}) can be formulated as
\begin{equation}
    Q_{a,t}(s,a,\omega)
    =
    Q_{f,t}(s,a,\omega)
    +
    K_{q,t} \big( r_{t+1}(s,a) - r_{f,t+1}(s,a,\omega) \big).
\end{equation}
At this stage, a new expression for the predicted reward $r_{f}(s,a,\omega)$ must be defined. 
As in classical temporal-difference methods, two distinct formulations arise, 
corresponding to the on-policy and off-policy settings. 
These yield approximate Bayesian counterparts of the SARSA and Q-learning 
algorithms, respectively \cite{Sutton1998}. The predicted reward can be written as
\begin{equation}
r_{f,t+1}(s,a,,\omega)
=
Q_{\pi,f,t}(s,a,\omega) - \gamma \, \mathbb{E}_\pi\big[ Q_{\pi,f,t+1}(S_{t+1}, A_{t+1},\omega) \,\big|\, S_t = s, A_t = a \big].
\end{equation}
The expectation over the next state and action can be approximated in two ways. 
In the SARSA approach, a sample of the expectation is taken by simulating the next action $a'$ according to a (possibly exploratory) policy $\pi$. 
Alternatively, in the Q-learning approach, the expectation is approximated using the greedy action under the current policy: $b = \arg \max_{a \in A} Q(s',a,\omega)$. 
Each of these methods effectively samples from a policy that may differ from the optimal policy, 
so the resulting value functions are not guaranteed to be identical. 
Accordingly, the forecast reward model can be written for GMKF-TD-SARSA and GMKF-TD-Q-learning as
\begin{eqnarray}
{r}_{f,t+1}(s,a,\omega)
&=& Q_{\pi,f,t}(s,a,\omega)
- \gamma Q_{\pi,f,t+1}(s',a',\omega)
- \gamma \varepsilon_t^{\text{SARSA}} \\
{r}_{f,t+1}(s,a,\omega)
&=& Q_{\pi,f,t}(s,a,\omega)
- \gamma \max_{b \in A} Q_{\pi,f,t+1}(s',b,\omega)
- \gamma \varepsilon_t^{Q}
\end{eqnarray}
where $\varepsilon_t^{\text{SARSA}},\varepsilon_t^{Q}$ are the corresponding sampling errors similar to the one given in Eq.~(\ref{eqn:epsilon_V}).

Once the Q-value function is estimated, one has to pick the optimal action. As our approach is probabilistic, and we get the posterior random variable describing $Q_{a,t}(s,a,\omega)$, the optimal action can be chosen in several ways. 
In $\epsilon$-greedy exploration \cite{Sutton1998}, the parameter $\epsilon$ specifies the probability of selecting a random action uniformly rather than following the current policy:
\begin{equation}
a =
\begin{cases}
\arg\max_a \, \mathbb{E}(Q_{a,t}(s,a,\omega)) & \text{with probability } 1 - \epsilon \\
\text{random action} & \text{with probability } \epsilon.
\end{cases}
\end{equation}
The $\epsilon$-greedy policy is accompanied by an exponentially decaying term to promote early exploration vs later exploitation of the value function. The $\epsilon$ term is therefore defined as
\begin{eqnarray}
\label{eqn:e_greedy}
    \epsilon_{k+1} = \max(\epsilon_\text{min}, \gamma_\epsilon \epsilon_k ) 
\end{eqnarray}
where $\epsilon_\text{min}$, $\epsilon_0$ and the decaying rate $\gamma_\epsilon$ are fine-tuned per application example. Another alternative exploration policy is the use the basis functions to maximize the weight of information during exploration \cite{Malekzadeh2020}.

However, the previously described strategies are based on the mean value of  the posterior Q-value function, and therefore do not take into account uncertainty being affiliated with the distribution.  The posterior distribution of $Q_{a,t}(s,a,\omega)$, $p_Q$, can be sampled for specific realization $\hat{\omega}$ such that the action can be selected according to
\begin{equation}
    a^* = \arg \max_a Q_{a,t}(s,a,\hat{\omega}),
\end{equation}
also known as Thompson sampling \cite{Thompson1933, Gupta2011}. This approach efficiently balances exploration and exploitation: action value functions with higher variance and higher mean are more likely to be selected. When the posterior variance decreases, the policy naturally becomes more greedy. The method removes the hyper-parameter tuning of an $\epsilon$-greedy policy.

Instead of sampling the posterior distribution, one might also introduce the confidence bounds and then select appropriate action. An exploration strategy in this direction is the Upper Confidence Bound (UCB) that does not explicitly model or sample from a full posterior distribution over the value function, but instead uses a compressed representation of uncertainty to guide action selection. Rather than maintaining an entire distribution over $Q_{\pi,a,t}(s,a,\omega)$, UCB relies on summary statistics such as an estimate of the mean $\mu(s,a)=\mathbb{E}(Q_{\pi,a,t}(s,a))$ and a measure of uncertainty (e.g., variance or a concentration-based confidence radius). The key idea is to construct an optimistic estimate of the value of each action by adding an exploration bonus to the mean, typically of the form 
\begin{equation}
  \mathcal{J}_Q(s,a)=  \mu(s,a) + \beta \sigma(s,a)
\end{equation}
in which $\sigma(s,a)$ is the standard deviation of $Q_{\pi,a,t}(s,a,\omega)$ and $\beta$ is the exploration coefficient. The action is then selected greedily with respect to this upper confidence estimate,
\[
a^* = \arg\max_a J_Q(s,a).
\]
While Gaussian assumptions can motivate the use of standard deviation as an uncertainty measure, UCB is more general and can be derived from non-parametric concentration bounds such as Hoeffding or Bernstein inequalities \cite{Seldin2011PAC-BayesianMartingales}. In contrast to Thompson sampling, which represents uncertainty through sampling entire plausible models from a posterior, UCB encodes uncertainty deterministically through upper bounds, providing a principled but more compressed form of exploration.

\section{Numerical examples}

This section presents several numerical examples to evaluate the proposed method. The initial examples serve as verification cases, where the performance of the method is assessed against reference solutions. The final example considers a more realistic application involving a partial differential equation-based model to demonstrate the applicability of the proposed approach to practical problems.

\subsection{Linear time invariant mass-spring-damping system}

We consider a discrete-time problem derived from a linear mass--spring--damper system. 
The system evolves according to
\begin{equation}
    s_{t+1} = A s_t + B a_t,
\end{equation}
where
\begin{equation}
    A =
    \begin{bmatrix}
        1 & \Delta t \\
        - k \Delta t & 1 - c \Delta t
    \end{bmatrix},
    \quad
    B =
    \begin{bmatrix}
        0 \\
        \Delta t
    \end{bmatrix}.
\end{equation}
The  state is given as
\begin{equation}
    s_t =
    \begin{bmatrix}
        x_t \\
        v_t
    \end{bmatrix}
    \in \mathbb{R}^2,
\end{equation}
where $x_t$ denotes position and $v_t$ velocity.
Here, $k > 0$ is the spring constant, $c \ge 0$ is the damping coefficient, and $\Delta t > 0$ is the time step. 

The goal is to maximize the expected discounted return for the reward defined as 
\begin{equation}
    r(s_t, a_t) = - \left(s_t^\top Q s_t + a_t^\top R a_t \right),
\end{equation}
with
\begin{equation}
    Q =
    \begin{bmatrix}
        1 & 0 \\
        0 & q_v
    \end{bmatrix},
    \quad
    R = r_a > 0.
\end{equation}


To obtain the reference solution for the optimal policy, we consider the LQR optimal control problem with the objective function same as reward only with positive sign \cite{Farjadnasab2022Model-freeLearning, Kiumarsi2014ReinforcementDynamics}.  Specifically, we assume that the optimal value function takes the quadratic form
\begin{equation}
V^*(s) = - s^\top P s,
\end{equation}
where \(P \succeq 0\) is a symmetric positive semidefinite matrix to be determined. This quadratic structure is consistent with linear-quadratic optimality problem \cite{Anderson1989OptimalMethods} and anticipates a linear  policy.

The Bellman optimality equation is given by
\begin{equation}
V^*(s) = \max_a \left[ - (s^\top Q s + a^\top R a) + \gamma V^*(A s + B a) \right].
\end{equation}
Substituting the quadratic ansatz one obtains
\begin{equation}
V^*(s) = \max_a \left[
- s^\top Q s - a^\top R a - \gamma (A s + B a)^\top P (A s + B a)
\right].
\end{equation}

Expanding the quadratic term in the dynamics gives
\begin{align}
(A s + B a)^\top P (A s + B a)
&= s^\top A^\top P A s + 2 a^\top B^\top P A s + a^\top B^\top P B a.
\end{align}

The objective is therefore a concave quadratic function in \(a\), and the first-order optimality condition yields
\begin{equation}
(R + \gamma B^\top P B)a^* = - \gamma B^\top P A s.
\end{equation}
Solving for the optimal action shows that the optimal policy is linear in the state,
\begin{equation}
a^*(s) = -K s,
\end{equation}
where the feedback gain matrix is
\begin{equation}
K = (R + \gamma B^\top P B)^{-1} (\gamma B^\top P A).
\end{equation}

The matrix \(P\) must satisfy a self-consistency condition obtained by substituting the optimal policy back into the Bellman equation. This leads to the discrete algebraic Riccati equation (DARE)
\begin{equation}
\label{riccati}
P = Q + \gamma A^\top P A
- \gamma^2 A^\top P B (R + \gamma B^\top P B)^{-1} B^\top P A.
\end{equation}

Finally, when the state is decomposed as \(s_t = (x_t, v_t)\), the optimal action can be expressed in component form as
\begin{equation}
a_t = -k_1 x_t - k_2 v_t,
\end{equation}
where the gain vector is given by
\begin{equation}
K = [k_1 \;\; k_2].
\end{equation}
This representation emphasizes that the optimal action is a linear state-feedback policy whose coefficients are determined implicitly through the Riccati equation.

To apply temporal difference type of approaches, we first need to parametrize the value function. Inline with the previous explanations, we choose the linear parametrization with quadratic kernel function in state:
\begin{equation}
V(s_t) \approx \theta^\top \phi(s_t),\quad 
\phi(s) =
\begin{bmatrix}
-x^2 \\
-x v \\
-v^2
\end{bmatrix}
\end{equation}
where the vector of parameters is learned from data via temporal difference updates rather than derived from the Riccati equation.
Hence, we choose
\begin{align}
    \Delta t &= 0.05, \\
    k &= 1, \\
    c &= 0.1, \\
    q_v &= 0.1, \\
    r_a &= 0.01, \\
    \gamma &= 0.95.
\end{align}

For space exploration, the states are initialized by a normal Gaussian distribution such that $s_0 \sim N(0,1)$, and simulated for 100 steps, or until the states reach  $x,y \leq 0.001$ \footnote{The termination update includes the next state value function. Early termination is only performed to avoid overfitting on the zero point parameters, and to speed up simulation time}. For the classical TD algorithm with $\alpha$ learning rate, the rate is set at a constant value of 0.1. For the GMKF-TD algorithm, the parameters are discretized using an ensemble of size 1000, and are initialized by a multivariate normal distribution of zero mean and covariance $I$, where $I$ is the identity matrix. As the transitions are deterministic, and so is the reward function, the measurement error variance is set to a small value of $10^{-2}$ to mimic real world conditions. For the updates of the parameters of the value function the stationary version of the filter is used, as shown in Eqs.~(\ref{eq:stationary_td_update})-(\ref{eq:stationary_kalman_gain}).
 
\begin{figure}
    \centering
    \includegraphics[trim={0 0 0 0},clip,width=0.45 \linewidth]{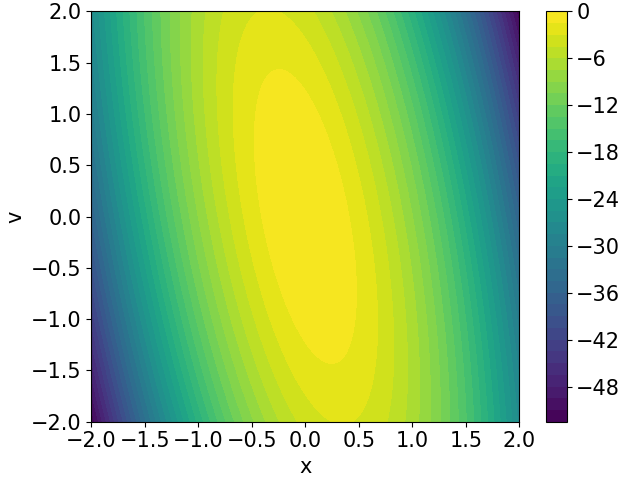}
    \caption{Value function for a range of positions and velocities, approximated by the GMKF-TD algorithm.}
    \label{fig:msd_det_V}
\end{figure}

The obtained posterior mean of the value function is plotted for a range of positions and velocities, as shown in Fig.~(\ref{fig:msd_det_V}) after training for 1000 episodes. As can be seen, the highest value (yellow region) is observed around $(x \approx 0, v \approx 0)$, corresponding to the stable equilibrium of the system and therefore a low-cost region. As the state moves away from the origin, the value function decreases smoothly, reflecting the increasing cumulative cost associated with larger deviations in position or velocity, which correspond to higher stored energy in the mass--spring--damper system. The level sets take the form of tilted ellipses, indicating a coupling between position and velocity induced by the system dynamics matrix $A$, which mixes the evolution of displacement and velocity over time.

\begin{figure}
    \centering
    \includegraphics[trim={0 0 0 0},clip,width=0.45 \linewidth]{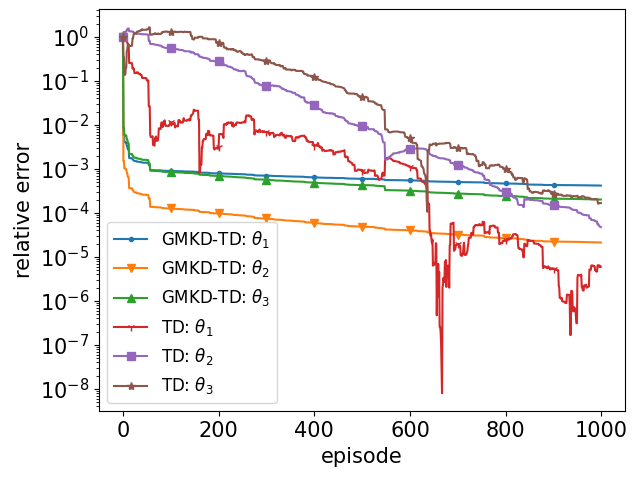}
    \includegraphics[trim={0 0 0 0},clip,width=0.45 \linewidth]{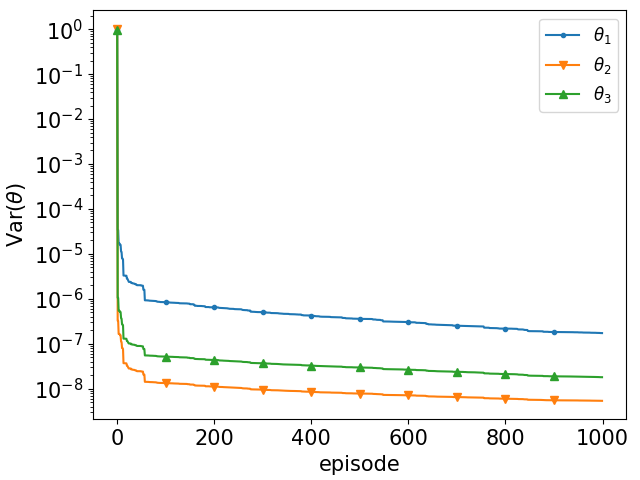}
    \caption{Relative error w.r.t. the DARE analytical solution and the two temporal difference algorithms (left). Variance of $\theta$ parameters after each episode using GMKF-TD with ensemble discretization (right).}
    \label{fig:msd_det_relerror}
\end{figure}

Solutions obtaned by the classical and GMKF-TD algorithms are compared to the reference value for the optimal value of $V$. 
This value is obtained by analytical solving of DARE. In our case, for the quadratic kernel function, unknown parameters $\theta$ correspond to the square matrix as $\theta_1:=P_{11}$, $\theta_2:=P_{12}+P_{21}$, and $\theta_3:=P_{22}$.
The relative error w.r.t. solution of the discrete Riccati equation is plotted for the parameters trained using the TD-$\alpha$ algorithm and GMKF-TD algorithm as shown in Fig.~(\ref{fig:msd_det_relerror}). All methods exhibit a decreasing trend, indicating convergence toward the reference value. The GMKD-TD variants for $\theta_1$, $\theta_2$, and $\theta_3$ demonstrate faster and more stable convergence compared to the standard TD baselines, achieving lower relative error within fewer episodes. In particular, GMKD-TD for $\theta_1$ shows the most rapid reduction, reaching near machine precision accuracy after approximately $600$ episodes, while $\theta_2$ and $\theta_3$ converge more gradually but still outperform their non-GMKD counterparts. In contrast, the standard TD methods show slower convergence and higher variance, with occasional fluctuations in the error trajectory. Overall, the results highlight the benefit of the GMKD-enhanced learning scheme in improving both convergence speed and accuracy relative to the reference solution.
 Additionally, the GMKF-TD approach brings additional information in terms of posterior variance. In Fig.~(\ref{fig:msd_det_relerror}) the posterior variance of each of parameters is plotted against episode.  All parameters exhibit a rapid initial decrease in variance, followed by a gradual convergence toward a low-variance regime. This indicates progressive stabilization of the learning process and increasing confidence in the estimated parameters. Among the three parameters, $\theta_2$ achieves the lowest variance, followed closely by $\theta_3$, while $\theta_1$ retains comparatively higher variance throughout training, suggesting slower convergence or higher sensitivity to stochastic updates. Overall, the results demonstrate that the learning algorithm effectively reduces uncertainty in the parameter estimates over time, leading to a stable asymptotic regime.  where the variance drops to a small value and slowly drifts towards zero after each episode.  

\subsection{Learning the Q-value function for the mass-spring-damper system}
For active control of the system, we can introduce an augmented state 
\begin{equation}
z =
\begin{bmatrix}
s \\
a
\end{bmatrix},
\end{equation}
where $Q(s,a) = -z^\top P_Q z$ is an exact quadratic form of the action-value function, for states $s$ and action $a$. The augmented matrix $P_Q$ can be analytically computed same as before, with the matrix being in the form 
\begin{equation}
P_Q =
\begin{bmatrix}
Q + \gamma A^\top P A \quad \gamma A^\top P B \\
\gamma B^\top P A \quad R + \gamma B^\top P B
\end{bmatrix},
\end{equation}
computed directly from the Riccati equation from Eq.~(\ref{riccati}). The action-value function can be therefore parameterized linearly using a quadratic kernel function in state and action:
\begin{equation}
Q(s_t,a_t) \approx \theta^\top \phi(s_t,a_t),\quad 
\phi(s,a) =
\begin{bmatrix}
-x^2 \\
-x v \\
-v^2 \\
-x a \\
-v a \\
- a^2
\end{bmatrix}
\end{equation}
The action-value function can be approximated in a continuous state and action space. For greedy action choices, the optimal action value function can be computed by $a^* = \arg \max(Q(s_t,a_t))$ leading to  $a^* = - (\theta_3 x + \theta_4v)/(2\theta_5)$.

The environment is initialized at state  $s_0 \sim N(0,1)$ and simulated for 100 steps or until the state reaches $x,y\leq 0.001$. An $\epsilon$-greedy exploration is used for training, with random exploration of the action in the uniform space $a \sim U(-20,20)$ with $\epsilon_0=1.0$, $\epsilon_\text{min}=0.001$, and decay term $\gamma_\epsilon=0.99$ (see Eq.~(\ref{eqn:e_greedy})).

\begin{figure}
    \centering
    \includegraphics[trim={0 0 0 0},clip,width=1 \linewidth]{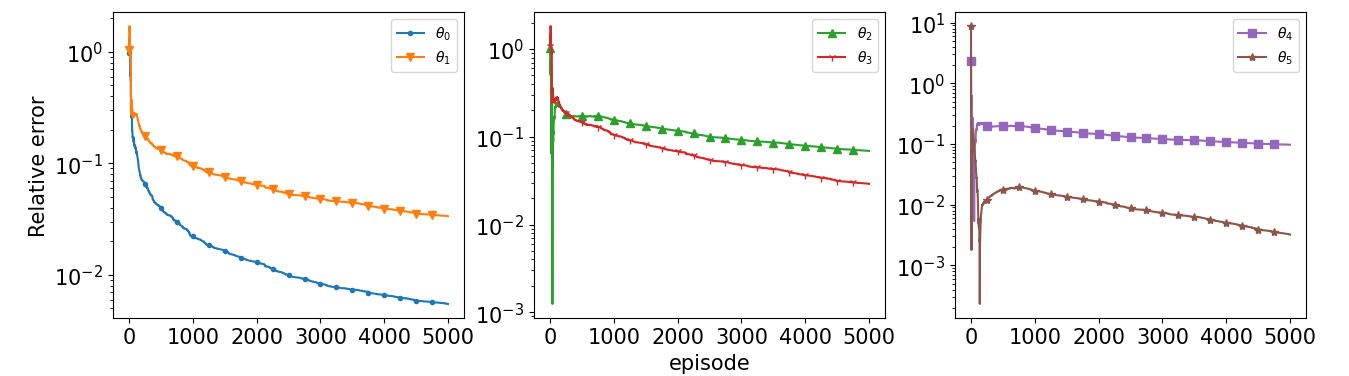}
    \caption{Relative error of parameters $\theta$ with analytical Riccati solution, including action parameters (left). }
    \label{fig:msd_det_theta_rel}
\end{figure}

\begin{figure}
    \centering
    \includegraphics[trim={0 0 0 0},clip,width=0.45 \linewidth]{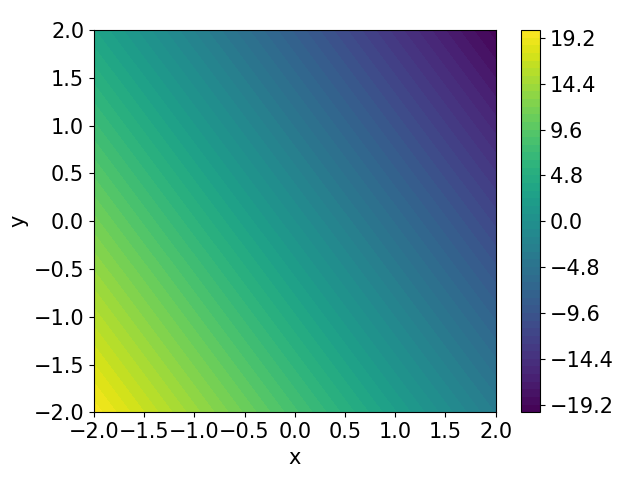}
    \caption{Learned policy function $\pi$, computed from the trained $Q$-value function using GMKF-TD algorithm for 5000 episodes.}
    \label{fig:msd_det_Pi}
\end{figure}

The GMKF-TD (SARSA) algorithm is used with ensemble discretization of the $Q$-value function. The ensemble is of size 100, with a zero mean prior $\mathbb{E}[\bm{\theta}_{f,0}]=\bm{0}$ and covariance $\text{Cov}(\bm{\theta}_{f,0})= I$, where $I$ is the identity matrix. The errors are modeled as $\eta \sim \mathcal{N}(0,0.01)$ and $\varepsilon_Q = \eta_Q \phi(s',a')$ where $\eta_Q\sim \mathcal{N}(0,0.001)$ (see Eq.~(\ref{main_update_Bellman_kf_final})). The algorithm is trained for 5000 episodes. The mean posterior value of the parameters is compared with the exact parameters derived by the Riccati equation. The relative error over 1000 episodes is shown in Fig.~(\ref{fig:msd_det_theta_rel}), for each parameter. The approximated parameters from the GMKF-TD algorithm closely converge to the analytical solution. The learned policy is shown in Fig.~(\ref{fig:msd_det_Pi}), after training. Based on the parameterization of the $Q$-value function, the policy solely depends on the last three parameters, while the first three terms represent the value function as in the previous section. 

\begin{figure}
    \centering
    \includegraphics[trim={0 0 0 0},clip,width=0.95 \linewidth]{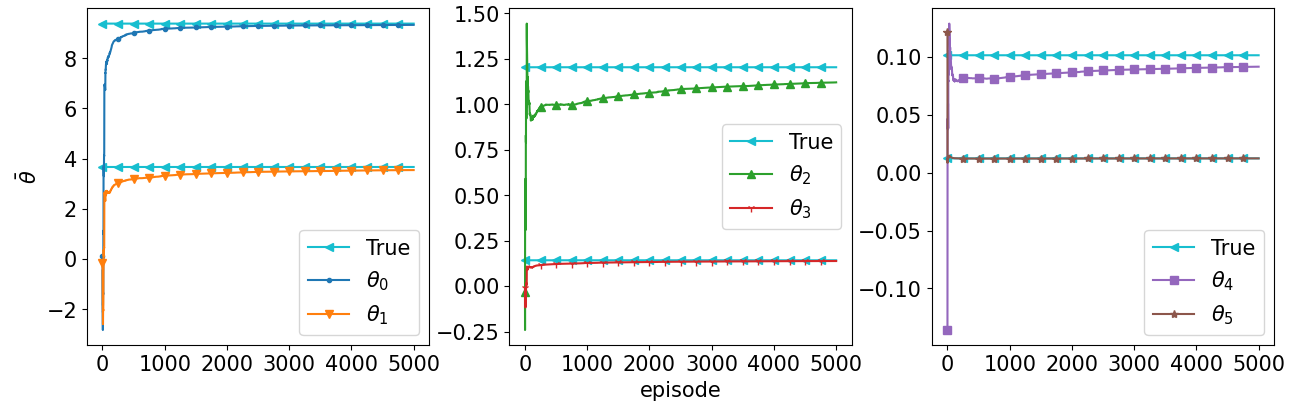}
    \caption{Mean of posterior parameters $\theta$, including action parameters, trained for 5000 episodes using an ensemble GMKF-TD algorithm.}
    \label{fig:msd_det_theta_mean}
\end{figure}

\begin{figure}
    \centering
    \includegraphics[trim={0 0 0 0},clip,width=1 \linewidth]{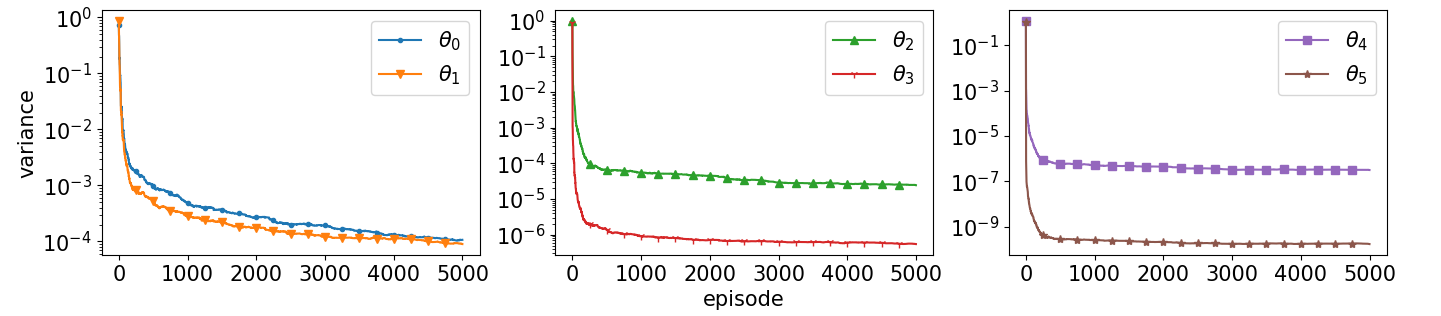}
    \caption{Variance of posterior parameters $\theta$, including action parameters, trained for 5000 episodes using an ensemble GMKF-TD algorithm.}
    \label{fig:msd_det_theta_var}
\end{figure}

\begin{figure}
    \centering
    \includegraphics[trim={0 0 0 0},clip,width=0.45 \linewidth]{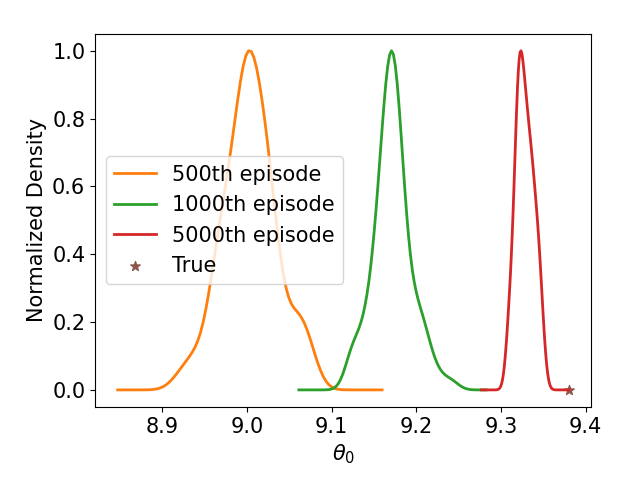}
    \caption{Probability density function of parameter $\theta_1$ at multiple episodes.}
    \label{fig:msd_det_theta_pdf}
\end{figure}

The current formulation of the $Q$-value function, and the quadratic kernel are not normalized, therefore the scale of the states and the actions are different, which impacts the variance of the parameters. The posterior mean of the parameters from the GMKF-TD algorithm can be seen in Fig.~(\ref{fig:msd_det_theta_mean}). The values move towards the true parameters in each episode. The rate is dependent on the choice of exploration rate and also the prior assumptions done for measurement and model error. Fig.~(\ref{fig:msd_det_theta_var}), shows the posterior variance of the same parameters after training for 5000 episodes. As there is no added noise to the sequential update of the parameters, the variance reduces towards zero with more observations.
Due to the measurement error and the sequential updating, the posterior mean slowly shifts towards the true mean with additional observations. Fig.~(\ref{fig:msd_det_theta_pdf}) shows the probability density function for the first parameter of the value function $\theta_0 x^2$ after 500, 1000, and 5000 episodes respectively. From the graph, the posterior slowly converges to a Gaussian-like distribution, with the mean being closer to the true mean value with more episodes. Here the exploration is almost at zero, which means that after 1000 episodes, the algorithm learns almost exclusively from a greedy policy. Note that here due to exloration strategy the posterior does not always contain the true value of the parameter. The reason for this is that due simplfiications the prior on the parameter is modelled as stationary while in reality we have exploration that shifts the prior by using random walk. Hence, the pdfs are behaving as shown. 

\subsection{Covariance estimation for linear time-invariant mass-spring-damper system}

We extend the previous implementation of the GMKF-TD algorithm to include covariance estimation as given in Eq.~(\ref{correction_eq}). We use a linear mapping of order 1, as in Eq.~(\ref{eq:low_triang}). The Frobenius norm (see Eq.~(\ref{eq:cov_min})) is minimized using BFGS optimization \cite{Liu1989OnOptimization}, and the square root factorization in Eq.~(\ref{correction_eq}) is computed using Cholesky decomposition. 

An ensemble of 1000 samples is used for the prior of the parameters $\theta$, same as previously. The GMKF-TD algorithms, with and without covariance correction, are trained under the same policy for 100 episodes. For comparison, after each iteration the prior is replaced with the posterior of the parameters as approximated by the GMKF-TD algorithm. 

\begin{figure}
    \centering
    \includegraphics[width=1\linewidth]{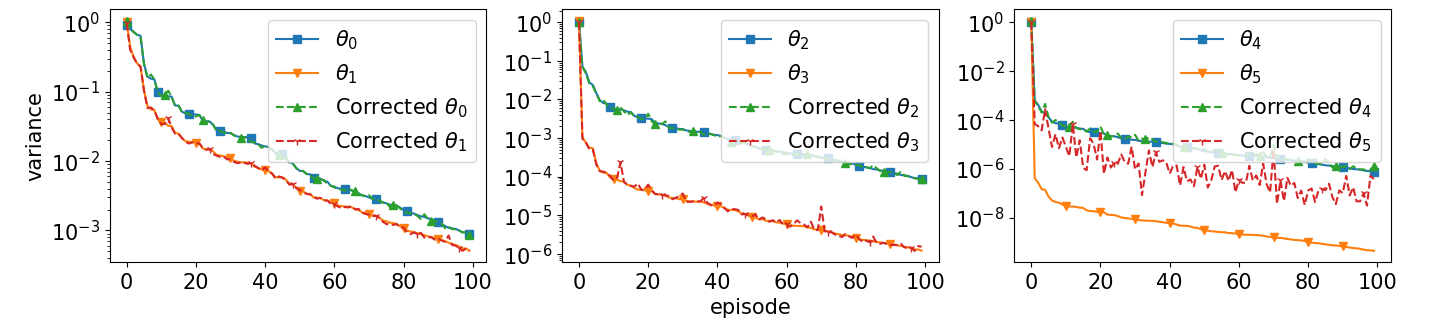}
    \caption{Variance of each parameter with and without the covariance correction update. Algorithms follow the same exploration strategy and parameter prior at each step.}
    \label{fig:msd_cc_var}
\end{figure}

In Fig.~(\ref{fig:msd_cc_var}), the posterior variance of the parameters is plotted for each parameter. The variance deviate in the two methods, once the values become smaller than the numerical errors. This is due to numerical approximations and ill-conditioning of the inverted matrix during covariance updating, see Eq~(\ref{correction_eq}). The variance of the 6th parameter (the one associated with the square action $-a^2$), drops below $10^{-6}$, at which point the posterior variance of the parameter differs between the two methods. Due to the variation in approximating the posterior variance, the posterior mean after each update will also differ.

\begin{figure}
    \centering
    \includegraphics[width=0.32\linewidth]{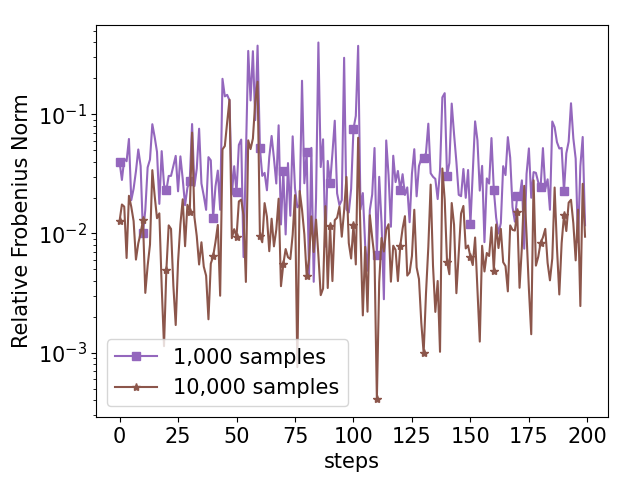}
    \includegraphics[width=0.32\linewidth]{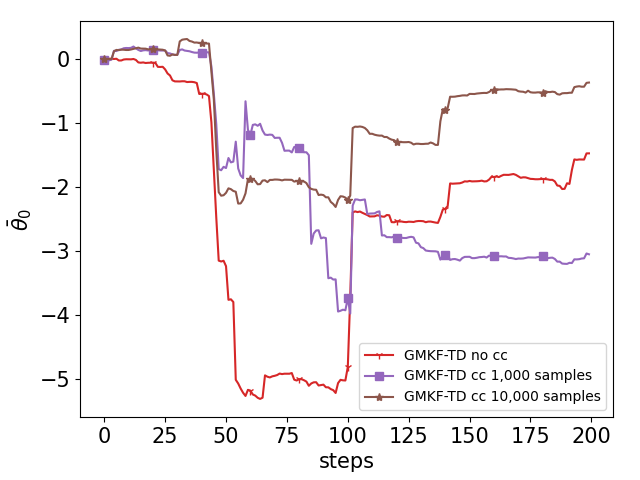}
    \includegraphics[width=0.32\linewidth]{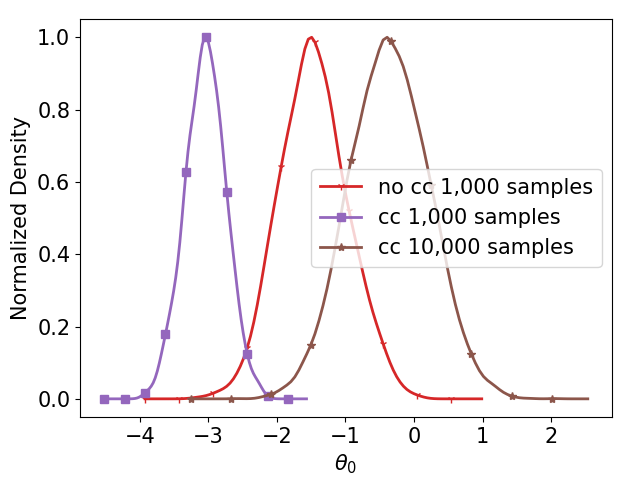}
    \caption{Frobenius relative norm error for two ensemble sizes (left) for first 200 steps. Mean value of first parameter after 200 steps using ensembles of different sizes with and without a correcting covariance update (middle). Probability density function of first parameter after 200 steps with varying ensemble size and covariance computation method (right).}
    \label{fig:msd_cc_fro}
\end{figure}

The fitness of the estimated covariance term is investigated using the GMKF-TD algorithm with covariance correction, before and after applying the correction update, see Eqs.~(\ref{main_update_Bellman_kf_final},\ref{main_gmkf_bel-time}). Note that we expect for this case that the covariance should be same, and hence this example serves as verification case. For this, the covariance matrices $C_1$ and $C_V$ are compared using the relative Frobenius norm equation 
\begin{equation}
    \text{relative error} = ||C_1 - C_V||_F/||C_1||_F
\end{equation}
where $C_1$ is the covariance computed by Eq.~(\ref{eq:cov_min}) and $C_V$ is the corrected covariance computed by Eq.~(\ref{correction_eq}). The covariance of the mean, $C_1$ is used to normalize the results. As the parameters are linear towards the value function and the prior is Gaussian, the two covariances should be approximately the same. Fig.~(\ref{fig:msd_cc_fro})-left shows the average of the relative Frobenius norm after each episode. The error denotes the difference between the two covariance matrices. As both matrices have numerical approximation errors, a deviation between the matrices is observed. While relatively small, these errors, added in sequential updating, can result in large deviations in the updating of the system. To partially alleviate this, one can use a larger sample-set in order to reduce the numerical approximation errors. Fig.~(\ref{fig:msd_cc_fro})-middle shows the update of two GMKF-TD algorithms with a correcting covariance step for parameter $\theta_1$, using an ensemble of 1000 samples, both updated independently on the same exploration strategy. The parameter $\theta_1$ is that associated with the squared position $x^2$ and its mean is plotted over 200 update steps. A slight improvement in the tracking of the mean can be observed, however the covariance matrix still suffers from numerical errors which are added up after each update. This is an indication that deviation between the two values is not only due to error approximation but also from numerical approximation. Fig.~(\ref{fig:msd_cc_fro})-right shows the PDF of the first parameter after 200 steps of the same algorithms. From the PDFs, we can see that the original $1,000$ ensemble with covariance correction underestimates the covariance, unlike the ensemble of $10,000$ samples that better matches the covariance of the distribution without the covariance correction update. 

Numerical approximations are expected due to the sequential updating nature of the algorithm, as the covariance tends to drive towards zero. When applicable, additional noise can be added apriori to the parameters to compensate for this, however depending on the design problem, this can affect learning. Numerical approximation errors are a result of small eigenvalues in the covariance, which in this example are prominant due to the scaling of the basis function $\Phi(s,a)$ caused by large values in the action, as seen in Fig.~(\ref{fig:msd_cc_var}). Normalization of the basis function can postpone and reduce the numerical approximation errors, when the parameters fitted are not near zero.

\subsection{Polynomial chaos expansion discretization for mass-spring-damper system} 

In the previous sections, we verified the numerical algorithm using the ensemble version of the proposed reinforcement learning algorithm. In problems with low number of random variables, we propose the use of polynomial chaos approximation instead of ensemble version. For the mass-spring-damper example, we approximate the six Q-value function parameters by a PCE with a Hermite polynomial basis. A polynomial order of 2 was chosen, which results in a total of $28$ coefficients \footnote{A polynomial order 1 is sufficient for linear and Gaussian systems, which reduces the total coefficients, however a larger coefficient is chosen for a comparison in the case of non-linear or non-Gaussian approximation}. The coefficients of prior that are different than zero are defined as shown in Tab.~(\ref{tab:PCE_coefficients}). 
\begin{table}[h]
\centering
\caption{Non-zero polynomial chaos expansion coefficients.}
\label{tab:PCE_coefficients}
\begin{tabular}{c c c}
\hline
Parameter & Multi-index $\boldsymbol{\alpha}$ & Coefficient \\ 
\hline
$\theta_0$ & $(1,0,0,0,0,0)$ & $\sqrt{0.8}$ \\
$\theta_0$ & $(2,0,0,0,0,0)$ & $\sqrt{0.1}$ \\
\hline
$\theta_1$ & $(0,1,0,0,0,0)$ & $\sqrt{0.8}$ \\
$\theta_1$ & $(0,2,0,0,0,0)$ & $\sqrt{0.1}$ \\
\hline
$\theta_2$ & $(0,0,1,0,0,0)$ & $\sqrt{0.8}$ \\
$\theta_2$ & $(0,0,2,0,0,0)$ & $\sqrt{0.1}$ \\
\hline
$\theta_3$ & $(0,0,0,1,0,0)$ & $\sqrt{0.8}$ \\
$\theta_3$ & $(0,0,0,2,0,0)$ & $\sqrt{0.1}$ \\
\hline
$\theta_4$ & $(0,0,0,0,1,0)$ & $\sqrt{0.8}$ \\
$\theta_4$ & $(0,0,0,0,2,0)$ & $\sqrt{0.1}$ \\
\hline
$\theta_5$ & $(0,0,0,0,0,1)$ & $\sqrt{0.8}$ \\
$\theta_5$ & $(0,0,0,0,0,2)$ & $\sqrt{0.1}$ \\
\hline
\end{tabular}
\end{table}

where multi-variate basis function is a Hermite polynomial basis with $\alpha$ being the multi-index. 

\begin{figure}
    \centering
    \includegraphics[width=1\linewidth]{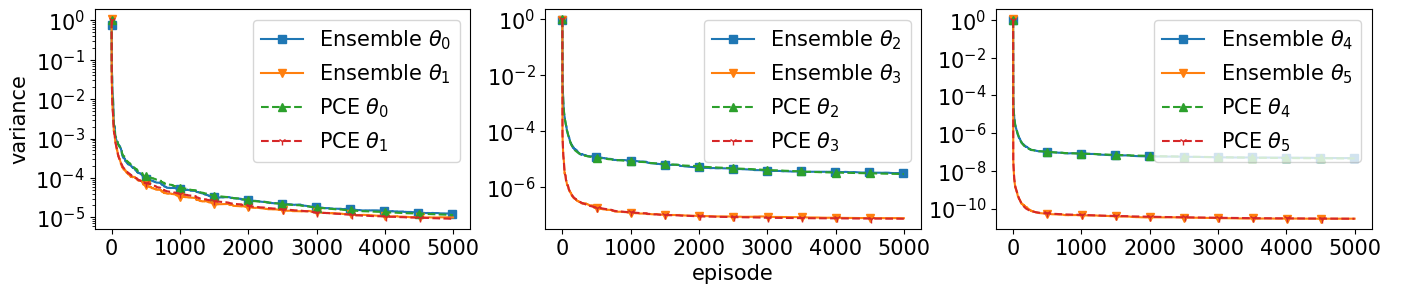}
    \caption{Posterior variance of parameters for mass-spring-damper including action parameters using PCE and ensemble discretization. }
    \label{fig:msd_pce}
\end{figure}

The algorithm is trained for 5000 episodes under the same exploration parameters and measurement error as in the previous section. The PCE approximation of the value function is computed in an algebraic manner .For comparison, an ensemble of 1000 samples is trained under the same exploration strategy and sampled from the previously defined prior distribution. For same exploration strategy, the greedy actions of the behaviour policy are selected by maximizing the value function approximated by the PCE coefficients. Fig.~(\ref{fig:msd_pce}) shows the posterior variance for each parameter using the ensemble and the PCE method. From the graph, the PCE and the ensemble approximate variance similarly with the variance reducing over more episodes and observations. This indicates that the ensemble size is sufficient to approximate the posterior distribution, similar to the PCE. As the PCE has a smaller number of coefficients (6x28 coefficients) compared to that of the ensemble (6x1000 samples), the computational time for the posterior is reduced \footnote{From our computations, the indicated efficiency is of factor 5 for the mass-spring-damper problem and PCE coefficients 6X28 compared to ensemble set of 6X1000.}.

\begin{figure}
    \centering
    \includegraphics[trim={0 0 3cm 0},clip,height = 4.9cm]{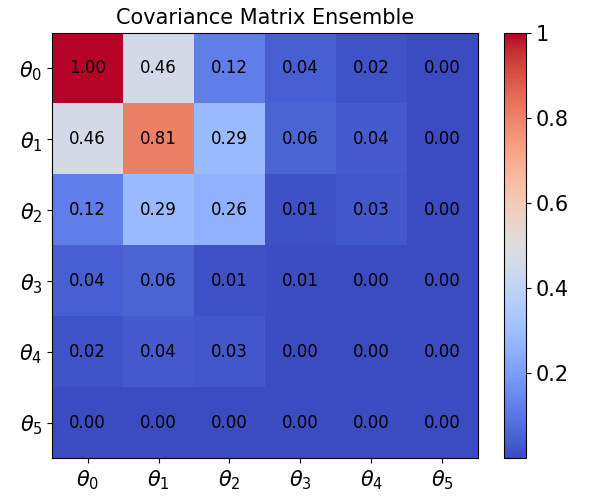}
    \includegraphics[height = 5cm]{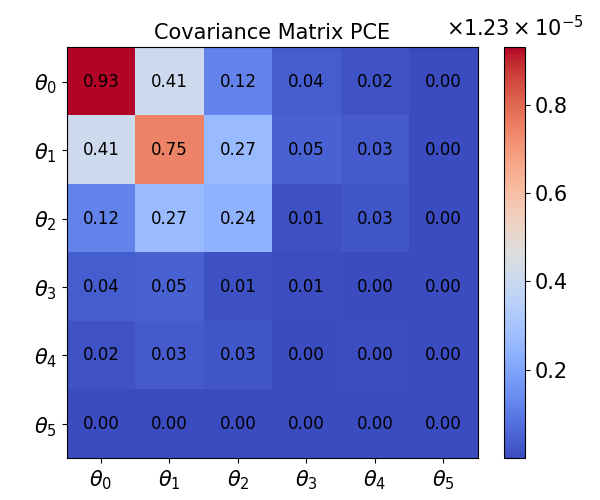}
    \caption{Covariance of posterior parameters approximated using an ensemble (left) and PCE (right) after 5000 episodes. }
    \label{fig:msd_pce_cov}
\end{figure}

Fig.~(\ref{fig:msd_pce_cov}) shows the full posterior covariance using both Ensemble and PCE approximations after training for 5000 episodes. From the plotted covariance matrices, there is a dependency across the parameters that was not defined in the prior. The covariance terms of parameters $\theta_4$ and $\theta_5$ are of smaller magnitude, which can also be seen by the variance from Fig.~(\ref{fig:msd_pce}).

\begin{figure}
    \centering
    \includegraphics[width=0.45\linewidth]{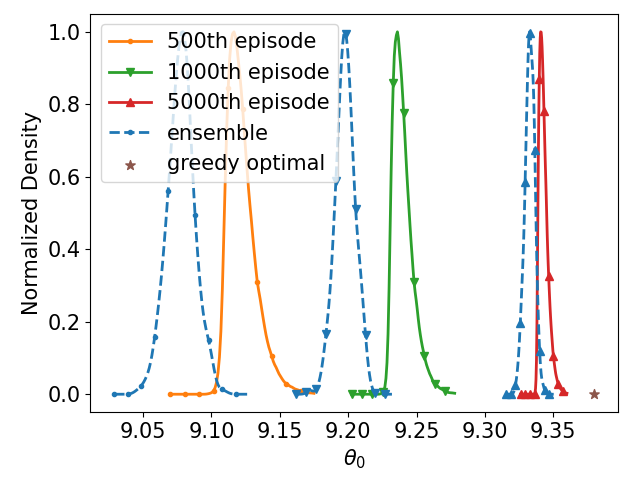}
    \caption{Probability density function for the first parameter $\theta_0$ at different episodes, using PCE discretization for exploration. }
    \label{fig:msd_pce_pdf}
\end{figure}

From Fig.~(\ref{fig:msd_pce_pdf}), the posterior probability function using both discretization methods is shown for episodes 500, 1000 and 5000 respectively. From the PDFs, the posterior mean of the PCE has a slightly better tracking of the optimal greedy value, as computed by the Riccati equation. The slight offset by the ensemble is caused by the exploration policy, as the ensemble is following the greedy exploration determined by the value function approximated with PCE coefficients. While the variance of the two methods is similar, previously shown in Fig.~(\ref{fig:msd_pce}), it does not overlap the greedy optimal value at each iteration. This is an effect caused by the behavior policy which includes exploration, as opposed to a greedy-only policy which is the one used to derive the true parameter values from Riccati equations. Riccati equations do not account for any noisy action selection, caused by the exploration. In other words, the GMKF-TD algorithm is approximating the posterior value function $V^{\tilde{\pi}}$ as opposed to the optimal value function $V^*$ \footnote{This is a non-stationary behavior which can be modeled by random walk, or similar stochastic processes and can be added in the aposterior parameter after each update}.  
\subsection{2D Differentially Heated Square Cavity}

\begin{figure}
    \centering
    \includegraphics[trim={0cm 0cm 10cm 0cm},clip,width=0.6\textwidth]{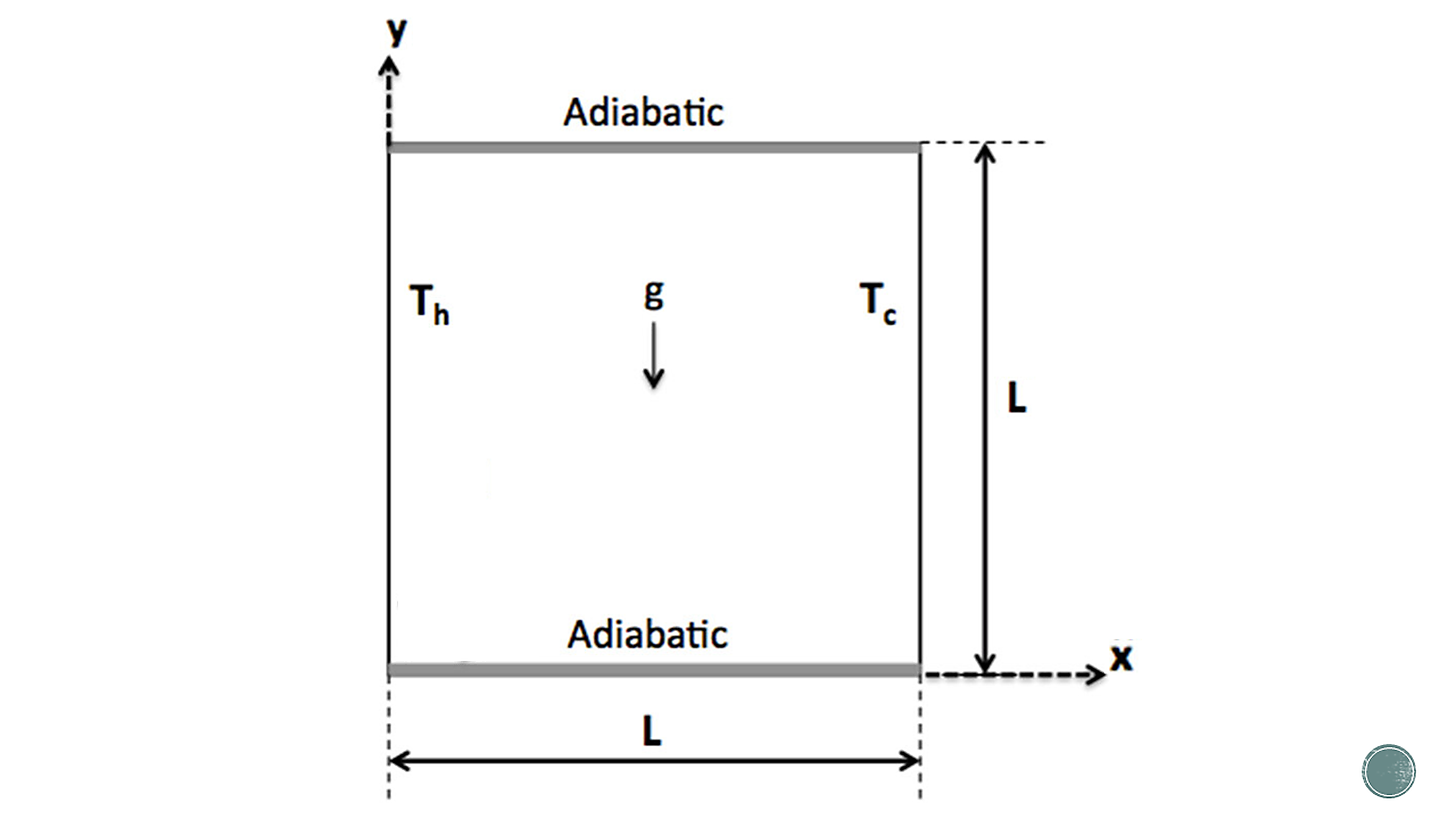}
    \caption{Schematic of 2D cavity with heated sidewalls}
    \label{fig:2dcavity}
\end{figure}

The two-dimensional differentially heated square cavity (see Fig.~\ref{fig:2dcavity}) is a classical benchmark for studying buoyancy-driven flows under the Boussinesq approximation \cite{Hachem2021DeepTransfer, Khanafer2003Buoyancy-drivenNanofluids}. 
The computational domain consists of a square cavity of side length $L$, filled with an incompressible Newtonian fluid. 
The left and right vertical walls are maintained at uniform but different temperatures, denoted $T_h$ and $T_c$, respectively ($T_h > T_c$), while the top and bottom horizontal walls are adiabatic. 
The resulting temperature difference $\Delta T = T_h - T_c$ induces density variations that drive natural convection within the enclosure.

Under the Boussinesq approximation, the flow is governed by the incompressible Navier--Stokes equations coupled with the energy equation:
\begin{align}
\nabla \cdot \mathbf{u} &= 0, \label{eq:continuity}\\[3pt]
\rho \left( \frac{\partial \mathbf{u}}{\partial t} + \mathbf{u}\cdot\nabla\mathbf{u} \right)
&= -\nabla p + \mu \nabla^2 \mathbf{u} + \rho g \beta (T - T_\mathrm{ref})\,\hat{\mathbf{y}}, \label{eq:momentum}\\[3pt]
\rho c_p \left( \frac{\partial T}{\partial t} + \mathbf{u}\cdot\nabla T \right)
&= k \nabla^2 T, \label{eq:energy}
\end{align}
where $\mathbf{u} = (u,v)$ is the velocity field, $p$ is the pressure, and $T$ is the temperature. 
The parameters $\rho$, $\mu$, $c_p$, and $k$ denote the density, dynamic viscosity, specific heat at constant pressure, and thermal conductivity, respectively. 
The coefficient $\beta$ is the thermal expansion coefficient, and $g$ is the gravitational acceleration acting in the vertical direction $y$. No-slip velocity conditions are applied on all boundaries (i.e, u=0 on all walls) and the thermal boundary conditions are imposed to the left and right wall.

The equations are made non-dimensional with the resulting Prandtl and the Rayleigh number
\begin{equation}
\Pr = \frac{\nu}{\alpha}, 
\qquad
Ra = \frac{g\,\beta\,\Delta T\,L^3}{\nu\,\alpha},
\end{equation}
with $\nu = \mu / \rho$ being the kinematic viscosity and $\alpha = k/( \rho c_p)$ being the thermal diffusivity.
In the present configuration, the target parameters are $\Pr = 0.71$ and $Ra \approx 10^4$. The parameters were defined as $\rho=37.89$, $c_p = 0.71$, $g=9.81$ and $\mu=k=\beta= \Delta T = L =1$.  

\begin{figure}
    \centering
    \includegraphics[trim={0 0cm 1.5cm 0.5cm},clip,height=5cm]{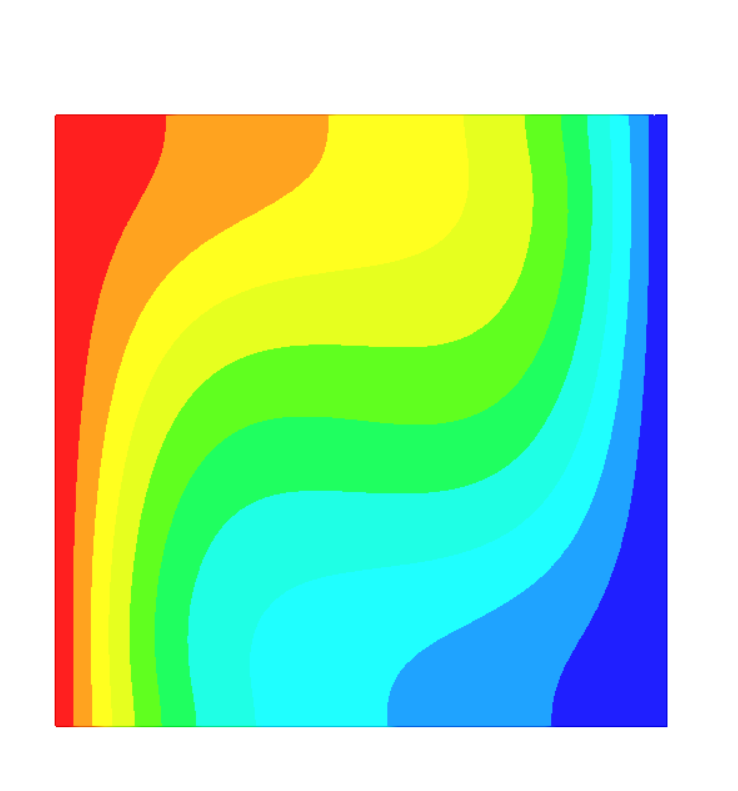}
    \includegraphics[trim={0cm 0cm 0 0},clip,height=5cm]{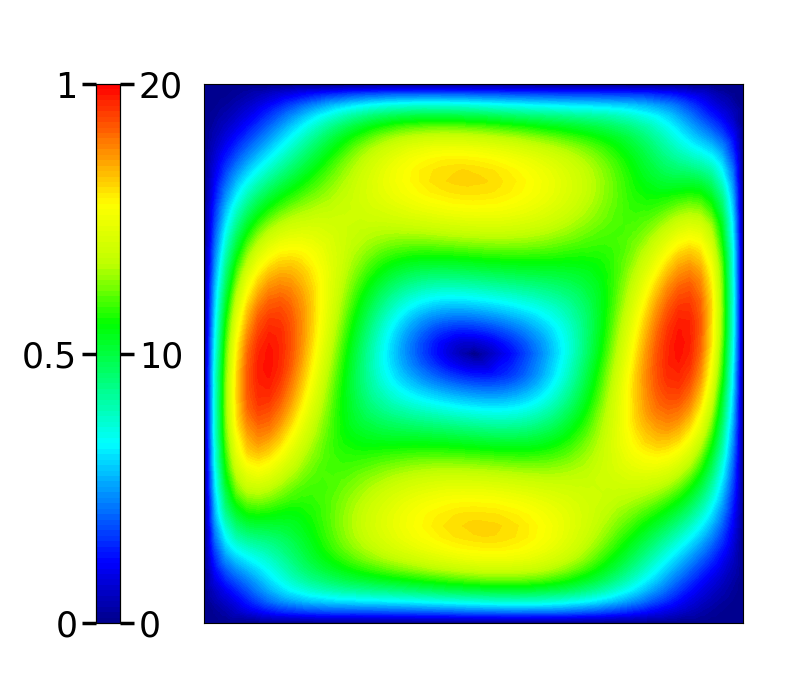}
    \caption{Steady-state of (Left) temperature and (Right) velocity in the closed cavity with uniformly heated walls.}
    \label{fig:cavity_nominal}
\end{figure}

The convective heat transfer at the hot wall is quantified using the Nusselt number. 
The local Nusselt number is defined as
\begin{equation}
Nu(y) = -\,\frac{L}{\Delta T}\,\left.\frac{\partial T}{\partial x}\right|_{x=0},
\end{equation}
and the mean (or average) Nusselt number is obtained by integrating along the heated wall
\begin{equation}
\overline{Nu} 
= \frac{1}{L}\int_0^L Nu(y)\,dy
= -\,\frac{1}{\Delta T}\int_0^1
\left(\frac{\partial T^*}{\partial x^*}\right)_{x^*=0} dy^*.
\end{equation}
In nondimensional form with $L=1$ and $\Delta T=1$, the average Nusselt number reduces to
\[
\overline{Nu} = -\int_0^1
\left(\frac{\partial T^*}{\partial x^*}\right)_{x^*=0} dy^*.
\]
For discretization of the 2D space, a uniform triangular meshing is used composed of 2500 nodes, with triangular elements. At steady state, as seen in Fig.~(\ref{fig:cavity_nominal}), the results show good agreement with the literature values reported in \cite{Hachem2021DeepTransfer}, with a computed Nusselt number of 2.23.

The aim is to control the temperature along the left wall in order to alleviate convective heat transfer, as discussed in \cite{Hachem2021DeepTransfer}. In the current configuration, the left wall is uniformly heated while the right wall is uniformly cooled. This creates convective heat that propagates over the closed cavity. An alternative heating configuration could uphold the same Rayleigh and Prandtl numbers while reducing the convective heat within the cavity. 

The system is described by a singular state which is similar to the actor only update of the single-step PPO algorithm shown in \cite{Hachem2021DeepTransfer} for the same benchmark problem. For this reason the notation for state is deducted from representations of action-value function such that instead of $Q(s,a)$, we are solving for $Q(s_0,a)=Q(a)$. This results in a time-independent problem, that is possible for $Ra <10^5$ where a stationary Nusselt number and steady flow can be reached \cite{Beintema2020ControllingLearning}. For time-dependent control, one can create a state space based on the cavity flow and temperature, using a 2D grid, and learn different control sequences at each timestep \cite{Beintema2020ControllingLearning}.

For the actions, we uniformly discretize the heated left wall into $n_l=10$ segments, each associated with an independent control input. Each segment is independently prescribed a heating temperature. We define an action $\pm \Delta T_\text{max}$ with $\Delta T_\text{max}=0.75$  applied over each segment of the wall, that defines the change of the temperature on that segment with respect to the default temperature $T_h=1$. To prevent a net temperature offset that would alter the global Rayleigh and Prandtl numbers, we scale and normalize the temperature at each segment. Let $\hat{T}_k:=a_k=\pm \Delta T_\text{max}$ denote the control temperature assigned to segment $k \in \{1,\ldots,n_l\}$. The control temperature is mapped to the actual wall temperature fluctuations $\tilde{T}_k$ after normalization and scaling by
\begin{equation}
\tilde{T}_k = 
\frac{\hat{T}_k - \langle \hat{\mathbf{T}} \rangle}{
\displaystyle \max\!\left\{ 1,\;
\frac{\max\limits_{\ell}|\hat{T}_\ell - \langle \hat{\mathbf{T}} \rangle|}{\Delta T_{\max}} \right\}},
\label{eq:T_scaling}
\end{equation}
where $\langle \cdot \rangle$ denotes the mean operator, in this case for the mean control temperature over all segments. The updated temperature per segment is computed using the existing heat temperature and the temperature fluctuations as
\beq
T_k = T_h + \tilde{T}_k
\eeq
where $T_h=1$ is the mean temperature on the left wall. The applied temperature fluctuations $\tilde{T}_k$ have zero spatial mean along the heated wall, such that $\langle \mathbf{T} \rangle = T_h$  which allows for segment based temperature control without impacting Rayleigh and Prandtl numbers. 

For the discrete action space
\[
\mathcal{A}=\{-\Delta T_{\max},+\Delta T_{\max}\},
\]
the action is represented through a binary feature mapping
\[
\Phi_a:\mathcal{A}\rightarrow\{0,1\}^{n_l}.
\]
For a given action \(a\), the resulting feature vector is denoted as
\[
\Phi_a(a)=
\left[
\ell_1(a),\ell_2(a),\ldots,\ell_{n_l}(a)
\right]^T,
\]
where $\ell_k(a)\in\{0,1\}$ denotes the action selected for the $k$-th segment. Specifically,
\begin{equation}
\ell_k(a)=
\begin{cases}
0, & \text{if } a_k=-\Delta T_{\max},\\
1, & \text{if } a_k=+\Delta T_{\max},
\end{cases}
\qquad k=1,\ldots,n_l.
\end{equation}
 The feature vector is augmented with a constant bias term to form the basis vector
\[
\Phi(a)=
\left[
\ell_1(a),\ell_2(a),\ldots,\ell_{n_l}(a),1
\right]^T .
\]
The action-value function is then parameterized linearly as
\[
Q(a)=\theta_{f,t}^{T}\Phi(a),
\]
where \(\theta_{f,t}\) contains the corresponding basis coefficients. The inclusion of the bias term results in a total of \(n_l+1=11\) parameters.

\begin{figure}
    \centering
    \includegraphics[trim={0 0 0.7cm 0},clip,width=5.05cm]{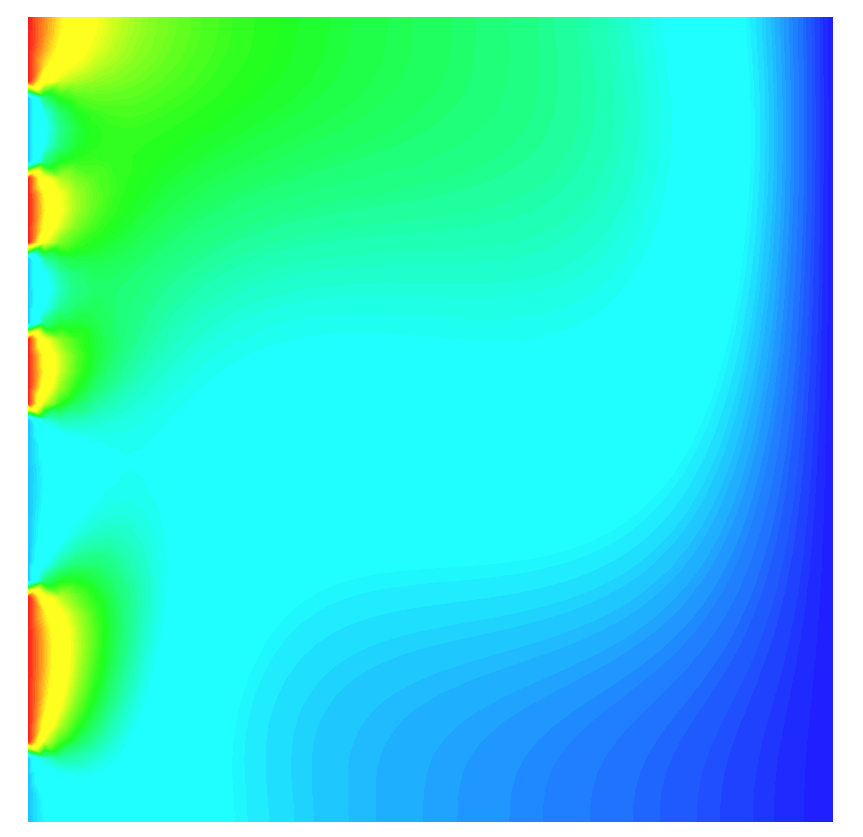}
    \includegraphics[trim={0cm 1cm 0 0},clip,width=.51\linewidth]{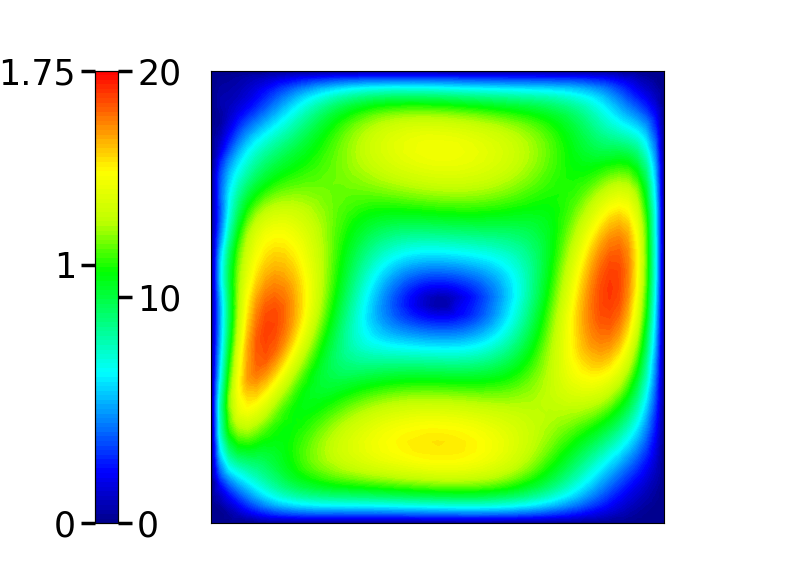}
    \caption{Steady-state of (Left) temperature and (Right) velocity in the closed cavity from the optimal state using 10 equidistant segments.}
    \label{fig:cavity_optimal}
\end{figure}

For each set of actions, the wall temperatures are computed and heat transfer is simulated until it reaches steady state flow ($t=20$). For reward, we use the negative average Nusselt number $r = -|Nu|$ at steady state flow. Due to a single state, the value function is directly updated by the observed reward (Nusselt number), resulting in a single-step horizon, with $\gamma=0$. In Figure \ref{fig:cavity_optimal} we show the optimal temperature distribution for temperature and velocity at steady-state over the closed cavity, for minimizing the Nusselt number, after simulating all possible combinations of segment heat activation. A lower Nusselt number indicates a reduction in the effectiveness of convective heat transfer, while convection itself persists, as evidenced by the roll-shaped flow patterns observed throughout the two-dimensional domain. 

\begin{figure}
    \centering
    \includegraphics[trim={0 0 0 0},clip,width=0.49 \linewidth]{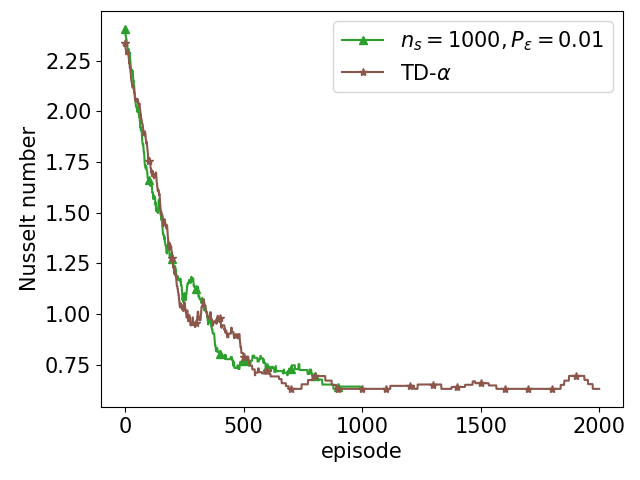}
    \includegraphics[trim={0 0 0 0},clip,width=0.49 \linewidth]{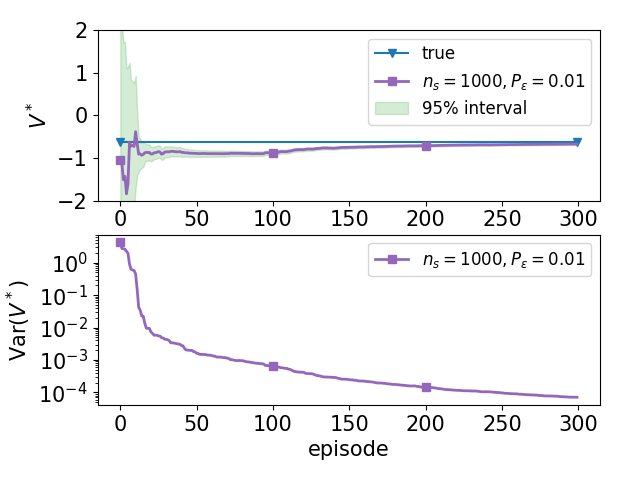}
    \caption{Plot of negative reward, i.e., absolute Nusselt number, for GMKF-TD algorithm and TD with decaying learning rate $\alpha$, over 2000 episodes (Left). Mean and variance of the value function at the optimal control temperature segment over 300 episodes (Right)}
    \label{fig:nusselt_td}
\end{figure}

The resulting environment, and consequently the value function, can be modeled as stationary, using the GMKF-TD algorithm from Eqs.~(\ref{eq:stationary_td_update}-\ref{eq:stationary_kalman_gain}). An $\epsilon$-greedy exploration policy is applied with a decaying term such that $\epsilon_0=1.0$, $\epsilon_\text{min}=0.01$, and decay term $\gamma_\epsilon=0.995$ (see Eq.~(\ref{eqn:e_greedy})). The TD algorithm (see Eq.~\ref{main_td}) with a decaying learning rate is also used. The learning rate is defined similar to exploration decay with parameters $\alpha_0=0.2$, $\alpha_\text{min}=0.01$, and decay term $\gamma_\alpha=0.995$. The learning rate parameters are chosen through trial and error, to achieve similar convergence pace with the GMKF-TD algorithm. 
The GMKF-TD algorithm is defined with zero-mean measurement error and variance $C_{\varepsilon_r}=0.01$ \footnote{The simulations of the Nusselt number, for the closed cavity example, are deterministic, however measurement error is present due to stochastic exploration policy}. The prior $\bm{\theta}_{f,0}$ is Gaussian and discretized using an ensemble of size 1000, with zero-mean and covariance $\text{cov}(\bm{\theta}_{f,0})= I$ where $I$ is the identity matrix. A total of 2000 episodes are run, using GMKF-TD algorithm and TD with decaying learning rate. In Fig.~(\ref{fig:nusselt_td})-left, the resulting Nusselt number is plotted over 2000 episodes, using both algorithms. From the graph, both algorithms converge to the same minimum value of ($\approx0.63$) which is the minimum Nusselt number achievable under the current formulation and meshing. The advantage of the GMKF-TD algorithm in this instance is that it replaced the learning rate with the Kalman gain update, without the need for trial and error. As the system is partially known, the parameter prior are given or can be chosen by experts which can be proven easier than fine-tuning the learning rate of the classical TD-$\alpha$ algorithm. Next to this, the variance information can help in creating adaptive exploration strategy. The GMKF-TD algorithm provides further insights to the variance of the value function. This can be seen from Fig.~(\ref{fig:nusselt_td})-right where the variance of the value function at the optimal control temperature is plotted. The Kalman update is acting as a sequential filter, reducing the variance towards zero after few hundred episodes, by sampling more observations. From the graph of the mean value function at the optimal temperature, the value function converges at optimal segment configuration (i.e., lowest Nusselt number) after 300 episodes, while the average Nusselt number per episode converges after 600 episodes when the exploration is minimized and transitions are primarily greedy. The discrepancy here is due to the $\epsilon$-greedy exploration, which continues to explore the environment with its own decaying rate. 

\begin{figure}
    \centering
    \includegraphics[trim={0 0 0 0},clip,width=0.45 \linewidth]{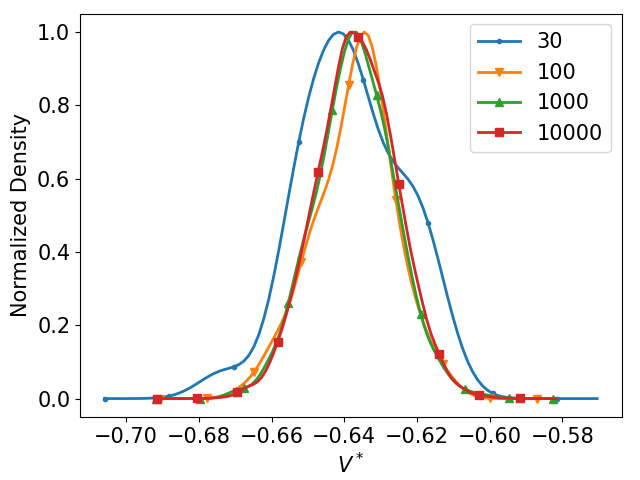}
    \includegraphics[trim={0 0 0 0},clip,width=0.45 \linewidth]{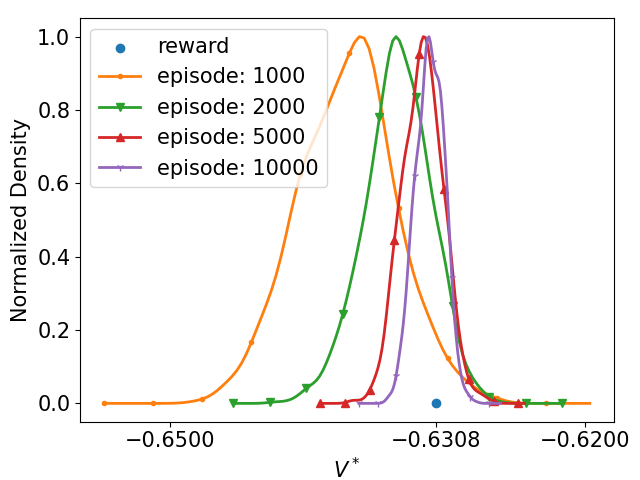}
    \caption{Probability density functions of the posterior value function at the optimal temperature state, approximated under different ensemble sizes (left). Probability density function of the value function at the optimal temperature state at different episodes (right).}
    \label{fig:cavity_ens_pdf_long}
\end{figure}

A further analysis is performed for the GMKF-TD algorithm, by modifying the ensemble size. Fig.~(\ref{fig:cavity_ens_pdf_long})-left illustrates the probability density function (PDF) of the posterior value function after 1,000 training episodes for ensemble sizes of 30, 100, 1,000, and 10,000 samples. In all cases, the posterior distributions are computed using the same set of observations per episode. With a Gaussian prior and linear parameter mapping of the value function, the posterior is expected to also be Gaussian. As shown in the figure, the posterior obtained with an ensemble size of 1,000 samples closely matches that of the 10,000-sample ensemble, indicating that the former provides a sufficiently accurate approximation of the underlying distribution. In contrast, the smaller ensemble sizes (30 and 100 samples) exhibit noticeably rougher approximations, as evidenced by deviations from the expected Gaussian shape of the PDF. 
The parameters are updated sequentially and can converge to the exact value, given that enough reward observations are collected, averaging over the measurement error and stochastic exploration policy. To visualize this, Fig.~(\ref{fig:cavity_ens_pdf_long})-right shows the PDF of the value function at the optimal control temperature, after 1000, 2000,5000, and 10000 episodes, using an ensemble of 1000 samples and with measurement error variance of $0.01$. From the pdfs, the mean slowly converges to the exact value of the observation, and the variance is progressively reduced, as the number of observations increase.

\begin{figure}
    \centering
    \includegraphics[trim={0 0 0 0},clip,width=0.45 \linewidth]{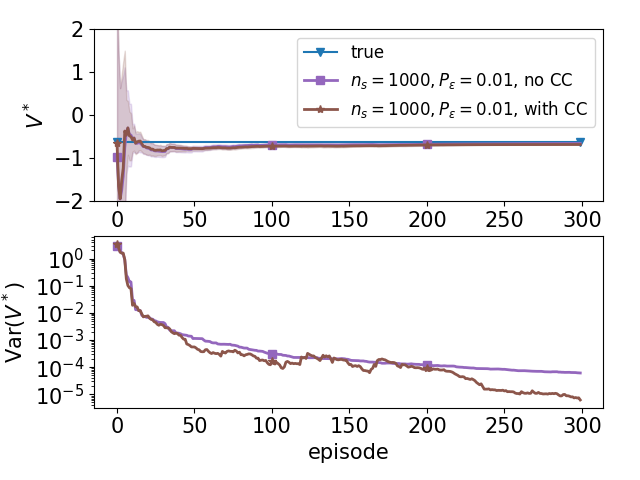}
    \includegraphics[trim={0 0 0 0},clip,width=0.45 \linewidth]{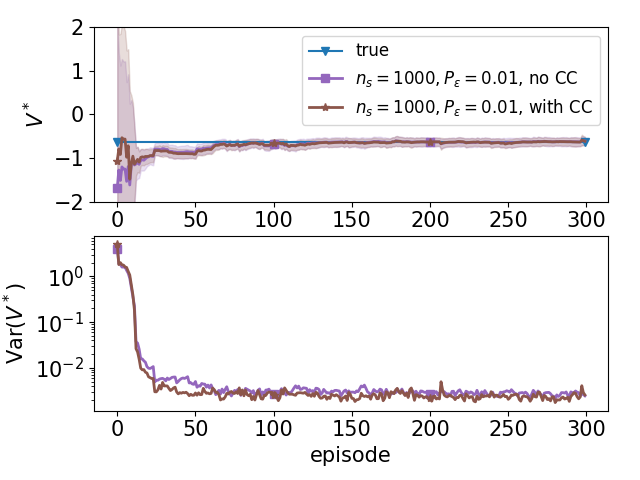}
    \caption{Mean and variance of value function at optimal control temperature segment, with and without a correction covariance step computed using the GMKF-TD algorithm. (Left) no added parameter noise. (Right) added parameter noise to prevent collapse of the covariance.}
    \label{fig:cavity_cc}
\end{figure}

With the ensemble size of 1000 samples and a measurement error of $0.01$, the GMKF-TD algorithm is expanded with a covariance correction update using Eq.~(\ref{corr_var}). From Fig.~(\ref{fig:cavity_cc}-left), the value function using the optimal control temperature segments is plotted, using both algorithms with and without the covariance correction step. The $\epsilon$-greedy exploration is performed using the parameters of the value function updated with the covariance correction step. Both algorithms converge towards the true value of the value function and the covariance is driven towards zero. Due to numerical approximations, the variance of the value function is fluctuating more over episodes. As shown also in the mass-spring-damper example, the accuracy of the covariance correction step drops, once the covariance becomes smaller than the numerical errors caused by the ensemble. A possible mitigation step is the addition of artificial noise apriori on the value function parameters, which prevents the covariance from being driven to zero. This is implemented in this example by adding a small noise in the parameters after each step such that $\theta_{f,t}=\theta_{f,t-1}+n_t$. The noise is modeled as a zero-mean Gaussian distribution with covariance computed by Cov$(n_t)=\eta \Phi(s_t)\odot I$ where $I$ is the identity matrix, multiplied element-wise with the basis function of the active segments of each step, and $\eta$ is a tuning parameter set at $1e^{-3}$. The presented noise model depends on the state, but not on the value function parameters, and allows for addition of noise without exploding the variance of inactive segments\footnote{Segments are referred to active or inactive due to their binary classification. When converted into temperature, the segments are either cold or hot with a $\pm \Delta T$ difference.}. Fig.~(\ref{fig:cavity_cc}-right) shows the tracking of the mean value for the value function and its variance for the optimal control temperature segments. As designed, the added noise prevents the variance from being driven to zero, and the behavior between the two algorithms aligns more. The implementation of the added noise in the value function parameters is not necessary for learning, and in this case it harms convergence towards the exact mean value without modeling any actual system dynamics. However it shows that for the case that the value function is linear and Gaussian, and the transitions are deterministic, the covariance computed using the Kalman gain is the correct covariance as validated by the covariance correction update. 

\newpage 
\pagebreak
\section{Conclusions}

In this work, we establish the equivalence between temporal-difference (TD) learning and the estimation of conditional expectations through projection-based formulations. Specifically, we show that the TD difference update can be interpreted as the innovation term of a probabilistic Kalman filtering framework, demonstrating that the classical TD update is recovered as a special case of the proposed Bayesian estimation approach. In contrast to classical ensemble filtering approaches, which primarily propagate the covariance associated with the estimated mean state or parameter distribution, the proposed framework explicitly estimates the conditional expectation of the second moment of the value function. This distinction is essential, as the conditional second moment is generally not equivalent to the covariance of the conditional mean; these quantities coincide only under restrictive assumptions, such as linear systems with Gaussian prior distributions. By incorporating higher-order probabilistic information, the proposed approach extends TD learning beyond deterministic value estimation toward a more general uncertainty-aware formulation capable of representing nonlinear and non-Gaussian systems.

For computational implementation, the uncertainty associated with the value function is parameterized using either ensemble-based sampling or polynomial chaos expansion (PCE). The ensemble formulation provides a direct Monte Carlo representation of the value function distribution, allowing the probabilistic Kalman updates to be performed using sampled realizations of the underlying uncertainty. Alternatively, PCE provides a more compact representation of stochastic variability by expressing the value function distribution through a finite set of polynomial basis coefficients and deterministic parameters, rather than requiring a large ensemble of samples. This formulation enables statistical quantities, including covariance and higher-order moments, to be computed directly from the expansion coefficients, thereby reducing errors associated with finite-sample covariance estimation. However, the applicability of PCE is limited to problems with moderate-dimensional stochastic parameterizations, where the resulting expansion remains computationally feasible.

The proposed GMKF-TD algorithm is evaluated on a range of mechanical and physics-based benchmark problems. Compared with conventional TD methods, the GMKF-TD framework introduces hyperparameters with a direct probabilistic interpretation, as parameter priors and uncertainty models can be derived from available knowledge of the underlying system. This reduces reliance on empirical tuning and provides a more systematic initialization of the learning process. Furthermore, the Kalman gain acts as an adaptive learning mechanism, automatically adjusting the magnitude of parameter updates according to the estimated uncertainty of the value function parameters and the reward model. This uncertainty-aware update mechanism can improve sampling efficiency, particularly in environments where observations are noisy or data are limited.

This work represents an initial step toward establishing a general probabilistic framework for temporal-difference learning with uncertainty propagation. Future research will focus on extending the proposed methodology to more complex and high-dimensional applications, as well as improving the numerical robustness of covariance correction and uncertainty propagation procedures, particularly in cases where distribution sampling and nonlinear transformations introduce additional approximation errors.

\bibliographystyle{elsarticle-num}
\bibliography{mybib}

@article{Sutton1998,
   author = {R.S. Sutton and A.G. Barto},
   doi = {10.1109/TNN.1998.712192},
   issn = {1045-9227},
   issue = {5},
   journal = {IEEE Transactions on Neural Networks},
   month = {9},
   pages = {1054-1054},
   title = {Reinforcement Learning: An Introduction},
   volume = {9},
   year = {1998},
}

@article{sutton1988td,
  author    = {Sutton, Richard S.},
  title     = {Learning to Predict by the Methods of Temporal Differences},
  journal   = {Machine Learning},
  volume    = {3},
  number    = {1},
  pages     = {9--44},
  year      = {1988},
  publisher = {Springer},
  doi       = {10.1007/BF00115009}
}

@book{Wiering2012,
   author = {Marco Wiering and Martijn van Otterlo},
   city = {Berlin, Heidelberg},
   doi = {10.1007/978-3-642-27645-3},
   isbn = {978-3-642-27644-6},
   publisher = {Springer Berlin Heidelberg},
   title = {Reinforcement Learning},
   volume = {12},
   year = {2012},
}

@article{Puterman2008,
   author = {Martin L. Puterman},
   doi = {10.1002/9780470316887},
   isbn = {9780470316887},
   journal = {Markov Decision Processes: Discrete Stochastic Dynamic Programming},
   month = {1},
   pages = {1-649},
   publisher = {wiley},
   title = {Markov decision processes: Discrete stochastic dynamic programming},
   year = {2008},
}

@article{Geist2010,
   author = {Matthieu Geist and Olivier Pietquin},
   journal = {Journal of Artificial Intelligence Research},
   pages = {483-532},
   title = {Kalman Temporal Differences},
   volume = {39},
   year = {2010},
}

@article{Geist2013,
   author = {Matthieu Geist and Olivier Pietquin},
   doi = {10.1109/TNNLS.2013.2247418},
   issn = {2162237X},
   issue = {6},
   journal = {IEEE Transactions on Neural Networks and Learning Systems},
   pages = {845-867},
   title = {Algorithmic survey of parametric value function approximation},
   volume = {24},
   year = {2013},
}

@book{Baird1995,
   author = {Leemon Baird},
   doi = {10.1016/B978-1-55860-377-6.50013-X},
   journal = {Machine Learning Proceedings 1995},
   pages = {30-37},
   publisher = {Elsevier},
   title = {Residual Algorithms: Reinforcement Learning with Function Approximation},
   year = {1995},
}

@inproceedings{Geerts2019, 
   series={CCN},
   title={Probabilistic Successor Representations with Kalman Temporal Differences},
   DOI={10.32470/ccn.2019.1323-0},
   booktitle={2019 Conference on Cognitive Computational Neuroscience},
   publisher={Cognitive Computational Neuroscience},
   pages = {399-402},
   author={Geerts, Jesse and Stachenfeld, Kimberly and Burgess, Neil},
   year={2019},
   collection={CCN} }

@article{Malekzadeh2020,
   author = {Parvin Malekzadeh and Mohammad Salimibeni and Arash Mohammadi and Akbar Assa and Konstantinos N. Plataniotis},
   doi = {10.1109/ACCESS.2020.3007951},
   issn = {21693536},
   journal = {IEEE Access},
   pages = {128716-128729},
   publisher = {Institute of Electrical and Electronics Engineers Inc.},
   title = {MM-KTD: Multiple Model Kalman Temporal Differences for Reinforcement Learning},
   volume = {8},
   year = {2020},
}

@article{Engel2003,
   author = {Yaakov Engel and Shie Mannor and Ron Melr},
   journal = {Proceedings of the 20th International Conference on Machine Learning (ICML-03)},
   title = {Bayes Meets Bellman: The Gaussian Process Approach to Temporal Difference Learning},
   year = {2003},
}

@inproceedings{Engel2005,
    author = {Engel, Yaakov and Mannor, Shie and Meir, Ron},
    title = {Reinforcement learning with Gaussian processes},
    year = {2005},
    isbn = {1595931805},
    publisher = {Association for Computing Machinery},
    address = {New York, NY, USA},
    doi = {10.1145/1102351.1102377},
    booktitle = {Proceedings of the 22nd International Conference on Machine Learning},
    pages = {201–208},
    numpages = {8},
    location = {Bonn, Germany},
    series = {ICML '05}
}

@misc{Lu2021,
      title={Robust and Adaptive Temporal-Difference Learning Using An Ensemble of Gaussian Processes}, 
      author={Qin Lu and Georgios B. Giannakis},
      year={2021},
      eprint={2112.00882},
      archivePrefix={arXiv},
      primaryClass={stat.ML},
}

@article{Bradtke1996,
   author = {Steven J. Bradtke and Andrew G. Barto},
   doi = {10.1007/BF00114723},
   issn = {0885-6125},
   issue = {1-3},
   journal = {Machine Learning},
   pages = {33-57},
   publisher = {Kluwer Academic Publishers},
   title = {Linear Least-Squares algorithms for temporal difference learning},
   volume = {22},
   year = {1996}
}

@article{Matthies2016,
   author = {Hermann G. Matthies and Elmar Zander and Bojana V. Rosić and Alexander Litvinenko},
   doi = {10.1186/S40323-016-0075-7},
   issn = {22137467},
   issue = {1},
   journal = {Advanced Modeling and Simulation in Engineering Sciences},
   month = {12},
   pages = {1-21},
   publisher = {Springer},
   title = {Parameter estimation via conditional expectation: a Bayesian inversion},
   volume = {3},
   year = {2016},
}

@article{Xiu2002,
   author = {Dongbin Xiu and George Em Karniadakis},
   doi = {10.1137/S1064827501387826},
   issn = {1064-8275},
   issue = {2},
   journal = {SIAM Journal on Scientific Computing},
   month = {1},
   pages = {619-644},
   title = {The Wiener--Askey Polynomial Chaos for Stochastic Differential Equations},
   volume = {24},
   year = {2002},
}

@article{Luenberger1968,
   author = {David G. Luenberger},
   journal = {Open Journal of Statistics},
   isbn = {978-0-471-18117-0},
   pages = {326},
   publisher = {Wiley},
   title = {Optimization by vector space methods},
   year = {1968},
}

@book{Kolmogorov1956,
   author = {Kolmogorov A.N.},
   edition = {Second English},
   publisher = {Chelsea Publishing Co.},
   title = {Foundations of the Theory of Probability },
   year = {1956},
}

@book{Bobrowski2005,
   author = {Adam Bobrowski},
   doi = {10.1017/CBO9780511614583},
   isbn = {9780521831666},
   month = {8},
   publisher = {Cambridge University Press},
   title = {Functional Analysis for Probability and Stochastic Processes},
   year = {2005},
}

@article{Lagoudakis2003,
   author = {Michail G Lagoudakis and Ronald Parr},
   journal = {Journal of Machine Learning Research},
   pages = {1107-1149},
   title = {Least-Squares Policy Iteration},
   volume = {4},
   year = {2003},
}

@article{Thompson1933,
   author = {William R. Thompson},
   doi = {10.2307/2332286},
   issn = {00063444},
   issue = {3/4},
   journal = {Biometrika},
   month = {12},
   pages = {285},
   publisher = {JSTOR},
   title = {On the Likelihood that One Unknown Probability Exceeds Another in View of the Evidence of Two Samples},
   volume = {25},
   year = {1933},
}

@article{Hayashi2020,
   author = {Kazuki Hayashi and Makoto Ohsaki},
   doi = {10.3389/FBUIL.2020.00059/BIBTEX},
   issn = {22973362},
   journal = {Frontiers in Built Environment},
   month = {4},
   pages = {514011},
   publisher = {Frontiers Media S.A.},
   title = {Reinforcement Learning and Graph Embedding for Binary Truss Topology Optimization Under Stress and Displacement Constraints},
   volume = {6},
   year = {2020},
}

@misc{Nguyen2021,
      title={Analytically Tractable Bayesian Deep Q-Learning}, 
      author={Luong Ha and Nguyen and James-A. Goulet},
      year={2021},
      eprint={2106.11086},
      archivePrefix={arXiv},
      primaryClass={cs.LG},
}

@article{Bojana2012,
   author = {Bojana V. Rosić and Alexander Litvinenko and Oliver Pajonk and Hermann G. Matthies},
   doi = {10.1016/J.JCP.2012.04.044},
   issn = {0021-9991},
   issue = {17},
   journal = {Journal of Computational Physics},
   month = {7},
   pages = {5761-5787},
   publisher = {Academic Press},
   title = {Sampling-free linear Bayesian update of polynomial chaos representations},
   volume = {231},
   year = {2012},
}

@article{Pajonk2013,
   author = {Oliver Pajonk and Bojana V. Rosić and Hermann G. Matthies},
   doi = {10.1016/J.CAGEO.2012.05.017},
   issn = {0098-3004},
   journal = {Computers \& Geosciences},
   month = {6},
   pages = {70-83},
   publisher = {Pergamon},
   title = {Sampling-free linear Bayesian updating of model state and parameters using a square root approach},
   volume = {55},
   year = {2013},
}

@article{Evensen2019,
   author = {Geir Evensen},
   doi = {10.1007/S10596-019-9819-Z/METRICS},
   issn = {15731499},
   issue = {4},
   journal = {Computational Geosciences},
   month = {8},
   pages = {761-775},
   publisher = {Springer International Publishing},
   title = {Accounting for model errors in iterative ensemble smoothers},
   volume = {23},
   year = {2019},
}

@article{Gupta2011,
   author = {Neha Gupta and Ole Christoffer Granmo and Ashok Agrawala},
   doi = {10.1109/ICMLA.2011.144},
   isbn = {9780769546070},
   journal = {Proceedings - 10th International Conference on Machine Learning and Applications, ICMLA 2011},
   pages = {484-489},
   title = {Thompson sampling for dynamic multi-armed bandits},
   volume = {1},
   year = {2011},
}

@article{Antos2008,
   author = {András Antos and Csaba Szepesvári and Rémi Munos},
   doi = {10.1007/S10994-007-5038-2/METRICS},
   issn = {08856125},
   issue = {1},
   journal = {Machine Learning},
   month = {4},
   pages = {89-129},
   publisher = {Springer},
   title = {Learning near-optimal policies with Bellman-residual minimization based fitted policy iteration and a single sample path},
   volume = {71},
   year = {2008},
}

@article{Salimibeni2020,
   author = {Mohammad Salimibeni and Parvin Malekzadeh and Arash Mohammadi and Konstantinos N. Plataniotis},
   doi = {10.1109/IEEECONF51394.2020.9443572},
   isbn = {9780738131269},
   issn = {10586393},
   journal = {Conference Record - Asilomar Conference on Signals, Systems and Computers},
   month = {11},
   pages = {579-583},
   publisher = {IEEE Computer Society},
   title = {Distributed Hybrid Kalman Temporal Differences for Reinforcement Learning},
   volume = {2020-November},
   year = {2020},
}

@article{Jang2022,
   author = {Seowoo Jang and Soyoung Yoo and Namwoo Kang},
   doi = {10.1016/j.cad.2022.103225},
   issn = {00104485},
   journal = {Computer-Aided Design},
   month = {5},
   pages = {103225},
   title = {Generative Design by Reinforcement Learning: Enhancing the Diversity of Topology Optimization Designs},
   volume = {146},
   year = {2022},
}

@article{Tziortziotis2017BayesianRegularization,
    title = {{Bayesian Inference for Least Squares Temporal Difference Regularization}},
    year = {2017},
    journal = {Lecture Notes in Computer Science (including subseries Lecture Notes in Artificial Intelligence and Lecture Notes in Bioinformatics)},
    author = {Tziortziotis, Nikolaos and Dimitrakakis, Christos},
    pages = {126--141},
    volume = {10535 LNAI},
    publisher = {Springer Verlag},
    isbn = {9783319712451},
    doi = {10.1007/978-3-319-71246-8{\_}8/FIGURES/4},
    issn = {16113349}
}

@article{Rosic2013ParameterSetting,
    title = {{Parameter identification in a probabilistic setting}},
    year = {2013},
    journal = {Engineering Structures},
    author = {Rosi{\'{c}}, Bojana V. and Ku{\v{c}}erov{\'{a}}, Anna and S{\'{y}}kora, Jan and Pajonk, Oliver and Litvinenko, Alexander and Matthies, Hermann G.},
    month = {5},
    pages = {179--196},
    volume = {50},
    doi = {10.1016/j.engstruct.2012.12.029},
    issn = {01410296}
}

@article{Siahkamari2020LearningDivergence,
    title = {{Learning to Approximate a Bregman Divergence}},
    year = {2020},
    journal = {Advances in Neural Information Processing Systems},
    author = {Siahkamari, Ali and XIA, XIDE and Saligrama, Venkatesh and Casta{\~{n}}{\'{o}}n, David and Kulis, Brian},
    pages = {3603--3612},
    volume = {33}
}

@article{Hachem2021DeepTransfer,
    title = {{Deep reinforcement learning for the control of conjugate heat transfer}},
    year = {2021},
    journal = {Journal of Computational Physics},
    author = {Hachem, E. and Ghraieb, H. and Viquerat, J. and Larcher, A. and Meliga, P.},
    month = {7},
    pages = {110317},
    volume = {436},
    publisher = {Academic Press},
    doi = {10.1016/J.JCP.2021.110317},
    issn = {0021-9991}
}

@article{Khanafer2003Buoyancy-drivenNanofluids,
    title = {{Buoyancy-driven heat transfer enhancement in a two-dimensional enclosure utilizing nanofluids}},
    year = {2003},
    journal = {International Journal of Heat and Mass Transfer},
    author = {Khanafer, Khalil and Vafai, Kambiz and Lightstone, Marilyn},
    number = {19},
    month = {9},
    pages = {3639--3653},
    volume = {46},
    publisher = {Pergamon},
    issn = {0017-9310}
}

@article{Kim2025AutomaticLearning,
    title = {{Automatic mesh generation for optimal CFD of a blade passage using deep reinforcement learning}},
    year = {2025},
    journal = {Journal of Computational Physics},
    author = {Kim, Innyoung and Chae, Jonghyun and You, Donghyun},
    month = {11},
    pages = {114306},
    volume = {541},
    publisher = {Academic Press},
    doi = {10.1016/J.JCP.2025.114306},
    issn = {0021-9991}
}

@article{Berry2016SemiparametricModels,
    title = {{Semiparametric modeling: Correcting low-dimensional model error in parametric models}},
    year = {2016},
    journal = {Journal of Computational Physics},
    author = {Berry, Tyrus and Harlim, John},
    month = {3},
    pages = {305--321},
    volume = {308},
    publisher = {Academic Press},
    doi = {10.1016/J.JCP.2015.12.043},
    issn = {0021-9991}
}

@misc{Ju2026SPDChallenges,
      title={SPD Matrix Learning for Neuroimaging Analysis: Perspectives, Methods, and Challenges}, 
      author={Ce Ju and Reinmar Kobler and Antoine Collas and Motoaki Kawanabe and Cuntai Guan and Bertrand Thirion},
      year={2026},
      eprint={2504.18882},
      archivePrefix={arXiv},
      primaryClass={cs.LG},
}

@article{Parish2024EmbeddedFasteners,
    title = {{Embedded symmetric positive semi-definite machine-learned elements for reduced-order modeling in finite-element simulations with application to threaded fasteners}},
    year = {2024},
    journal = {Computational Mechanics 2024 74:6},
    author = {Parish, Eric and Lindsay, Payton and Shelton, Timothy and Mersch, John},
    number = {6},
    month = {5},
    pages = {1357--1381},
    volume = {74},
    publisher = {Springer},
    doi = {10.1007/S00466-024-02481-5},
    issn = {1432-0924},
    arxivId = {2307.05434}
}

@article{Vishny2024High-DimensionalSamples,
    title = {{High-Dimensional Covariance Estimation From a Small Number of Samples}},
    year = {2024},
    journal = {Journal of Advances in Modeling Earth Systems},
    author = {Vishny, David and Morzfeld, Matthias and Gwirtz, Kyle and Bach, Eviatar and Dunbar, Oliver R.A. and Hodyss, Daniel},
    number = {9},
    month = {9},
    pages = {e2024MS004417},
    volume = {16},
    publisher = {John Wiley and Sons Inc},
    doi = {10.1029/2024MS004417;WGROUP:STRING:PUBLICATION},
    issn = {19422466}
}

@article{Schweitzer1985GeneralizedProcesses,
    title = {{Generalized polynomial approximations in Markovian decision processes}},
    year = {1985},
    journal = {Journal of Mathematical Analysis and Applications},
    author = {Schweitzer, Paul J. and Seidmann, Abraham},
    number = {2},
    month = {9},
    pages = {568--582},
    volume = {110},
    publisher = {Academic Press},
    doi = {10.1016/0022-247X(85)90317-8},
    issn = {0022-247X}
}

@article{Amherst2008ValueBasis,
    title = {{Value Function Approximation in Reinforcement Learning using the Fourier Basis}},
    year = {2008},
    journal = {Computer Science Department Faculty Publication Series},
    author = {Amherst, Scholarworks@umass and Konidaris, George and Osentoski, Sarah},
    volume = {101}
}

@article{Liu2025Model-freeChaos,
    title = {{Model-free robust reinforcement learning via Polynomial Chaos}},
    year = {2025},
    journal = {Knowledge-Based Systems},
    author = {Liu, Jianxiang and Wu, Faguo and Zhang, Xiao},
    month = {1},
    pages = {112783},
    volume = {309},
    publisher = {Elsevier},
    doi = {10.1016/J.KNOSYS.2024.112783},
    issn = {0950-7051}
}

@article{Zarrouki2024AdaptiveControl,
    title = {{Adaptive Stochastic Nonlinear Model Predictive Control with Look-ahead Deep Reinforcement Learning for Autonomous Vehicle Motion Control}},
    year = {2024},
    journal = {IEEE International Conference on Intelligent Robots and Systems},
    author = {Zarrouki, Baha and Wang, Chenyang and Betz, Johannes},
    pages = {12726--12733},
    publisher = {Institute of Electrical and Electronics Engineers Inc.},
    isbn = {9798350377705},
    doi = {10.1109/IROS58592.2024.10801876},
    issn = {21530866},
    arxivId = {2311.04303}
}

@misc{Lillicrap2015,
      title={Continuous control with deep reinforcement learning}, 
      author={Timothy P. Lillicrap and Jonathan J. Hunt and Alexander Pritzel and Nicolas Heess and Tom Erez and Yuval Tassa and David Silver and Daan Wierstra},
      year={2019},
      eprint={1509.02971},
      archivePrefix={arXiv},
      primaryClass={cs.LG},
}

@misc{Mnih2013PlayingLearning,
      title={Playing Atari with Deep Reinforcement Learning}, 
      author={Volodymyr Mnih and Koray Kavukcuoglu and David Silver and Alex Graves and Ioannis Antonoglou and Daan Wierstra and Martin Riedmiller},
      year={2013},
      eprint={1312.5602},
      archivePrefix={arXiv},
      primaryClass={cs.LG}, 
}

@misc{Haarnoja2018,
      title={Soft Actor-Critic: Off-Policy Maximum Entropy Deep Reinforcement Learning with a Stochastic Actor}, 
      author={Tuomas Haarnoja and Aurick Zhou and Pieter Abbeel and Sergey Levine},
      year={2018},
      eprint={1801.01290},
      archivePrefix={arXiv},
      primaryClass={cs.LG},
}

@article{Beintema2020ControllingLearning,
    title = {{Controlling Rayleigh–B{\'{e}}nard convection via reinforcement learning}},
    year = {2020},
    journal = {Journal of Turbulence},
    author = {Beintema, Gerben and Corbetta, Alessandro and Biferale, Luca and Toschi, Federico},
    number = {9-10},
    month = {10},
    pages = {585--605},
    volume = {21},
    publisher = {Taylor {\&} Francis},
    doi = {10.1080/14685248.2020.1797059},
    issn = {14685248},
    arxivId = {2003.14358}
}

@article{Farjadnasab2022Model-freeLearning,
    title = {{Model-free LQR design by Q-function learning}},
    year = {2022},
    journal = {Automatica},
    author = {Farjadnasab, Milad and Babazadeh, Maryam},
    month = {3},
    pages = {110060},
    volume = {137},
    publisher = {Pergamon},
    doi = {10.1016/J.AUTOMATICA.2021.110060},
    issn = {0005-1098}
}

@article{Kiumarsi2014ReinforcementDynamics,
    title = {{Reinforcement Q-learning for optimal tracking control of linear discrete-time systems with unknown dynamics}},
    year = {2014},
    journal = {Automatica},
    author = {Kiumarsi, Bahare and Lewis, Frank L. and Modares, Hamidreza and Karimpour, Ali and Naghibi-Sistani, Mohammad Bagher},
    number = {4},
    month = {4},
    pages = {1167--1175},
    volume = {50},
    publisher = {Pergamon},
    doi = {10.1016/J.AUTOMATICA.2014.02.015},
    issn = {0005-1098}
}

@article{Liu1989OnOptimization,
    title = {{On the limited memory BFGS method for large scale optimization}},
    year = {1989},
    journal = {Mathematical Programming},
    author = {Liu, Dong C. and Nocedal, Jorge},
    number = {1-3},
    month = {8},
    pages = {503--528},
    volume = {45},
    publisher = {Springer-Verlag},
    doi = {10.1007/BF01589116/METRICS},
    issn = {00255610}
}

@article{Seldin2011PAC-BayesianMartingales,
    title = {{PAC-Bayesian Inequalities for Martingales}},
    year = {2011},
    journal = {IEEE Transactions on Information Theory},
    author = {Seldin, Yevgeny and Laviolette, François and Cesa-Bianchi, Nicolò and Shawe-Taylor, John and Auer, Peter},
    number = {12},
    month = {10},
    pages = {7086--7093},
    volume = {58},
    doi = {10.1109/TIT.2012.2211334},
    issn = {00189448},
    arxivId = {1110.6886}
}

@book{Anderson1989OptimalMethods,
    title = {{Optimal Control: Linear Quadratic Methods}},
    year = {1989},
    booktitle = {Book},
    author = {Anderson, B. D. O. and Moore, J. B.},
    publisher = {Prentice-Hall},
    isbn = {0-13-638560-5}
}
\end{document}